\documentclass[runningheads]{llncs}
\usepackage{graphicx}
\usepackage{tikz}
\usepackage{comment}
\usepackage{amsmath,amssymb} % define this before the line numbering.
\usepackage{color}

\usepackage{float} 
\usepackage{hyperref}
\usepackage{amsmath}
\usepackage{amsfonts}
\usepackage{mathtools}
\usepackage{siunitx}
\usepackage{bbm}
\usepackage{svg} %https://tex.stackexchange.com/questions/299/how-to-get-long-texttt-sections-to-break
\usepackage[font=small,labelfont=bf]{caption}  % For figure captions https://tex.stackexchange.com/a/86121
\usepackage[titletoc]{appendix}% https://ctan.org/pkg/appendix, https://tex.stackexchange.com/a/44971

\newcommand{\Mod}[1]{\mathrm{mod} #1} % https://tex.stackexchange.com/a/137076

\begin{document}

% \renewcommand\thelinenumber{\color[rgb]{0.2,0.5,0.8}\normalfont\sffamily\scriptsize\arabic{linenumber}\color[rgb]{0,0,0}}
% \renewcommand\makeLineNumber {\hss\thelinenumber\ \hspace{6mm} \rlap{\hskip\textwidth\ \hspace{6.5mm}\thelinenumber}}
% \linenumbers
\pagestyle{headings}
\mainmatter
\def\ECCVSubNumber{1}

\title{Scene relighting with illumination estimation in the latent space on an encoder-decoder scheme} % Replace with your title

% INITIAL SUBMISSION 
\begin{comment}
\titlerunning{ECCV-20 submission ID \ECCVSubNumber} 
\authorrunning{ECCV-20 submission ID \ECCVSubNumber} 
\author{Alexandre Pierre Dherse, Martin Nicolas Everaert, Jakub Jan Gwizda{\l}a}
\institute{Paper version \ECCVSubNumber}
\end{comment}
%******************

% CAMERA READY SUBMISSION
%\begin{comment}
\titlerunning{Scene relighting with illumination estimation in the latent space}
% If the paper title is too long for the running head, you can set
% an abbreviated paper title here
%
\author{Alexandre Pierre Dherse\inst{1} \and
Martin Nicolas Everaert\inst{1} \and
Jakub Jan Gwizda{\l}a\inst{1}}
\authorrunning{A. Dherse, M. Everaert, J. Gwizda{\l}a}
% First names are abbreviated in the running head.
% If there are more than two authors, 'et al.' is used.
%
\institute{EPFL, Lausanne, Switzerland\footnote{CS-413 project supervised by Majed El Helou in class taught by Prof. Sabine S{\"u}sstrunk}\\
\email{\{alexandre.dherse,martin.everaert,jakub.gwizdala\}@epfl.ch}
}
%\end{comment}
%******************
\maketitle

\begin{abstract}
The image relighting task of transferring illumination conditions between two images offers an interesting and difficult challenge with potential applications in photography, cinematography and computer graphics. In this report we present methods that we tried to achieve that goal. Our models are trained on a rendered dataset of artificial locations with varied scene content, light source location and color temperature. With this dataset, we used a network with illumination estimation component aiming to infer and replace light conditions in the latent space representation of the concerned scenes.
\end{abstract}

\begin{figure}[H]
    \centering
    \begin{minipage}[c]{0.22\textwidth}
        \includegraphics[width=1\textwidth]{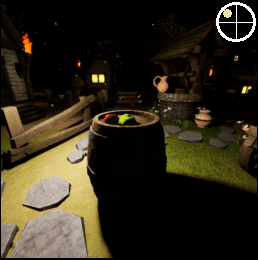}
    \end{minipage} \hfill
    \begin{minipage}[c]{0.22\textwidth}
        \includegraphics[width=1\textwidth]{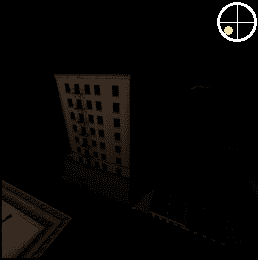}
    \end{minipage}\hfill
    \begin{minipage}[c]{0.22\textwidth}
        \includegraphics[width=1\textwidth]{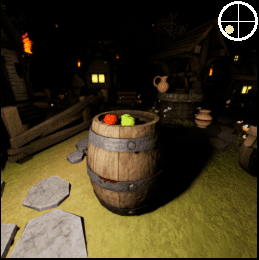}
    \end{minipage}\hfill
    \begin{minipage}[c]{0.22\textwidth}
        \includegraphics[width=1\textwidth]{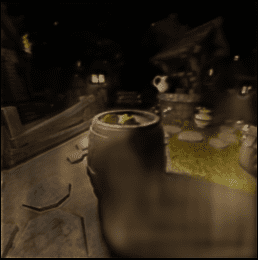}
    \end{minipage}
    \caption{Example of image relighting. From left to right: input $I$, target $T$ and ground-truth $G(I,T)$ images from the evaluation set; relit image $\hat{G}(I, T)$ produced by the model with illumination predicter (see \ref{ssec:illuminationPredicter}). The idea of having tiny diagrams in the upper right corner indicating the position of the main light source to facilitate interpretation comes from \cite{nodeeplearning}.}
    \label{fig:cover_illustration}
\end{figure}

\section{Introduction}

%\textbf{What is the problem in general terms?} 
In general terms, the relighting problem consists of taking two different scenes, with possibly different illuminations, and transferring the illumination from one scene to another. In the dataset there are \num{300} different scenes, each rendered with \num{5} different color temperatures and \num{8} different light directions, for a total of \num{12000} images \cite{vidit}. Some scenes are rendered from the same location (same objects) but with a different camera view.

%\textbf{Why is it important/of interest?} 
The solution to the described problem can be valuable for a few reasons. Firstly, it would be a useful tool to make our photos more beautiful by changing the lightning in already captured photographs. It is known that that professional photography may require advanced equipment to provide desired lighting conditions for shooting and that casual users are usually left with the illumination in a particular environment that they cannot affect \cite{vidit}. This method could also be used to perform data augmentation before training a particular network, allowing to achieve a wider variety in the dataset and potentially resulting in a model that is more stable under light modifications, i.e. less biased to specific conditions.

%\subsection{Related works}

%\textbf{What other solutions are available?}
% TO DO: develop on the first sentences, they must be more than using the general terms
Some related works are considering intrinsic image decomposition \cite{intrinsicwild,learningintrinsic}, photography enhancement \cite{underexposed}, style transfer \cite{cyclegan,stargan}, depth estimation \cite{depth} and shading annotation \cite{shading}. In \cite{taskonomy}, interesting relations are shown between some tasks, such as depth estimation, that can be performed on images.

%\textbf{Why are they not good enough? What would we ideally want to have?} 
On one hand, intrinsic image decomposition consists of separating image reflectance from shadings. Roughly speaking, reflectance are the ``real'' colors of the objects, independent from illumination, and shading is how illumination affects the reflectance to produce the final image. While it is useful for some application such as enhancing photography exposure, it does not modify the illumination. However, it is possible for neural networks to learn scene geometry elements and lighting, for instance to perform depth estimation and shading annotation \cite{depth,shading}. On the other hand, style transfer has shown really good results in modifying the style of an image in real-time, transforming for instance a video of horses into the same video where horses are replaced by zebras, performing the same actions \cite{cgan}. Ideally, we would like to be able to change the illumination -- shading -- of an input image to fit the one of a target image, while keeping the scene -- objects, reflectance -- of the input image.

%\subsection{Similar works}
%\textbf{Narrowing down the similar solutions, what are these good similar solutions?}
Image relighting is currently performed either with several input images of the same scene \cite{sparse}, or for specific tasks such as portrait relighting \cite{portrait,deepportrait} or daytime translation \cite{daytime}. Also note that not all those works extract the target illumination from a target image. The training dataset collection is often a bottleneck to train such networks if one wants to apply the image relighting to real sceneries -- it is particularly time consuming and technically difficult to collect photographs of multiple locations under different lighting conditions. However, there exist innovative ways of acquiring real-life datasets \cite{daytime,multiillumination}. The alternative is to use artificially produced datasets, such as the VIDIT dataset \cite{vidit} in this project.

%\textbf{How do they work?}
The image relighting networks trained in the papers above follow a similar schema. The input image is passed through an encoder to extract portrayed scene content information \cite{sparse,daytime,deepportrait} and/or illumination information -- environment map, lighting feature \cite{deepportrait,portrait}. Some operations occur in the encoded -- latent -- space, to add the target illumination -- target style, target lighting feature, target light. The modified latent space is then passed through a decoder to output the relit image with the input image's scene. In addition to the relit-image-generator (encoder-decoder) the papers \cite{deepportrait,daytime} also have a discriminator as the ones from style transfer \cite{cyclegan,stargan} to improve relit image quality and assure their realism.

%\textbf{What are their limitations?}
One limitation of those networks is that they are too task specific -- e.g. only work well on a portraits, a daytime change, 5 inputs of the same scene. We want to have something more general that would work for every scene and illumination. We also want it to extract the target illumination automatically from the target image.

%\subsection{Our proposition}
%\textbf{What are we proposing?}
We present algorithms constituting minor modifications to \cite{portrait} that take as input an image to be relit and a target image. By processing each of them, we aim to separate the illumination information from the scene information, either in the latent space only, or in the latent space and in the skip links. Then, the scene information from the input image and the illumination information from the target image are combined and decoded to produce a relit image.

%\textbf{How does it contribute/improve?}
We would like for our methods not to be specific to a particular content -- we hope that VIDIT dataset \cite{vidit} containing scenes with various objects would allow achieving that goal. Our solution does not require the target illumination to be explicitly given. Indeed, the target illumination information is extracted automatically from the target image. It also extracts the scene and illumination information from only one sample of each image -- input and target.

%\textbf{What do we base our work on, briefly how do we solve? }
The two images are processed though the same convolutional encoder to extract the scene and illumination information. We then concatenate the scene information from the input image and the illumination information from the target image and process them through a convolutional decoder to output a relit image. As we know in advance the scene and illumination, we also have at our disposal the ground-truth image this method is expected to output. We optimize the similarity between the relit image and the ground-truth image by iteratively updating the parameters of the encoder and the decoder.

\begin{figure}
    \centering
    \includegraphics[width=0.75\textwidth]{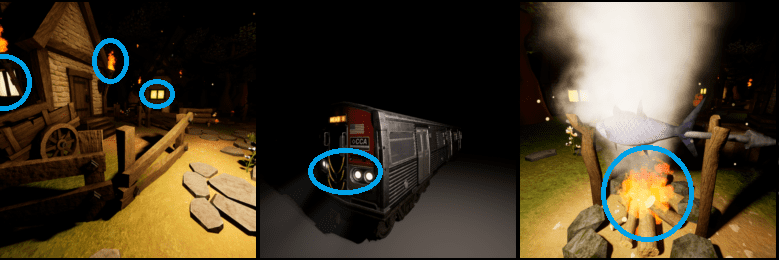}
    \caption{Non-main light sources in the VIDIT dataset \cite{vidit}.}
    \label{fig:lightsources}
\end{figure}

%\subsection{Assumptions}
%\textbf{What are the assumptions we make?}
In the VIDIT dataset \cite{vidit} we used to train our models, color temperatures and light directions are discrete. We assume it would be possible to generalize our method to continuous values. In this dataset, only the main source of light is changing direction and not the other light sources -- e.g. light from windows, subway train headlights, bonfire --~as the ones depicted in the figure \ref{fig:lightsources}. Therefore, those other light sources are not considered as illumination information by our method.

%\textbf{What will be presented about the method?}
In the following parts of the report, we present our literature review followed by the explanation of the architectures of the networks we tried. We describe different variations we have been experimenting with, together with the losses used for training.

\section{Literature review}
\subsection{Intrinsic image decomposition}

The intrinsic image decomposition problem consists of taking an image $I$ and decomposing it into two separate layers, a reflectance layer $R$ representing the geometry of the objects and a shading layer $S$ that depends on the lighting, with $I$ = $R \times S$. \cite{intrinsicwild} contains a dataset of human-labeled decomposition of images, and \cite{learningintrinsic} uses a convolutional neural network to obtain the decomposition, obtaining great results. The training of the network is done using unlabeled videos, with a fixed viewpoint but changing lighting.

\subsection{Image modification}

Style transfer is the process of changing a particular characteristic of an image to adopt the style of another one, for example adding color to a black and white image. \cite{img2img} provides a way to make these modifications while keeping the image realistic by jointly training a discriminator that learns to differentiate between real and generated data and a generator that tries to fool the discriminator. An encoder-decoder with skip-links is used for the generator, and a ``PatchGAN'' classifier for the discriminator. This discriminator only penalizes high-frequency artefacts, as it classifies each $N \times N$ patch of an image. This method requires to have an available dataset of the wanted style transfer. \cite{cyclegan} extends this strategy to unpaired mappings, allowing for example the transformation of a photograph to a painting, and \cite{stargan} generalizes the GAN by enabling it to learn mappings from and to multiple domains. An application of those techniques can be seen in \cite{daytime}, where the lighting of an image is changed to emulate any time of the day. 

Instead of learning an image-to-image translation, another way to meaningfully modify an image is by learning an image-to-illumination mapping. This mapping can then be used to relight underexposed photos for instance \cite{underexposed}.

\subsection{Relighting images}

While relighting a scene captured under a large number of different illuminations is an easy process, doing so with one or very few samples is a complex challenge, as multi-illumination datasets are hard to capture. \cite{multiillumination} developed a system to quickly capture scenes under several lighting conditions using a flash bouncing off the walls to illuminate the scene in different directions. It provides a dataset of real scenes captured under 25 lighting conditions, that can be used to train a relighting network. \cite{sparse} trains a fully convolutional encoder-decoder to relight a scene from a sparse set of input images. It needs 5 images of a same scene under different directional lighting to produce a relit image in a novel light.

Both \cite{portrait} and \cite{deepportrait} attempt to relight a portrait from a single image. \cite{portrait} uses a U-Net, an encoder-decoder with skip connections, to predict the illumination corresponding to the source image and replace it with the target lighting. The network produces great results on faces but the objective functions are applied only to masked part of the images -- people -- while the backgrounds are not affected but replaced by the blurred version of the environment map. \cite{deepportrait} uses a similar architecture, but the encoder extracts facial features in addition to lighting features. The facial feature stays unchanged while the target lighting replaces the input lighting. A PatchGAN is used to improve the quality of the generated image and removes local artifacts. This network gives good results on portrait images but does not generalizes to other scenes, giving poor results on non-portrait images.

\subsection{Other related tasks}
The paper \cite{taskonomy} aims to identify the relation between different tasks -- such as reshading, semantic segmentation, surface normals estimation. By using in one task the encoder trained in another task, they compute a similarity between the two tasks. They have for instance shown that reshading is close to finding occlusion edges, both of which are close to Z-depth and distance estimation, but very far from 2D-segmentation, camera pose estimation or scene classification.
% TODO: relate to our project or do not use it %Done? ~Martin
The paper \cite{depth} aims to produce a network for estimating distance between the camera and the points of the images. 
%Their data is only videos from a moving car on the street and they don't have ground-truths. Thanks to the network that estimates the distance and another one that estimates the camera pose, their loss roughly exploits the consistency between the evolution of the estimated camera pose and distances.
As it would have been the case if we would have kept the 1-to-1 transfer method\footnote{In our first experiments to get used to the dataset, we trained a slightly modified version -- output 3 channels instead of 1 -- of their depth estimation network (not using their camera pose network) to perform 1-to-1 relighting where inputs are $1024 \times 1024$ images with light coming from east and color temperature 2500\si{\kelvin} and ground-truths are $1024 \times 1024$ images with light coming from east and color temperature 6500\si{\kelvin}. Results of this familiarizing experiment are shown in the appendix, figure \ref{fig:beginningExperiment1to1}.}, their depth estimation network extracts information from one image -- the photograph of the street -- to produce a new image -- the depth map.

Authors of \cite{semantic} analyze how the manipulation of the latent representation on different depth levels of the generative network influences the semantical properties of the generated scenes. Even though it does not directly address the problem of image relighting, the paper claims that the generative networks learn to divide scene synthesis into a hierarchy of steps and that the representations responsible for the color scheme of the image are localized in the deepest part of the network.

\section{Implementation}

In \ref{ssec:dataset} the details about the dataset used for training and evaluation are specified. Section \ref{ssec:autoencoder} describe the architecture of the core parts of the network, shared between evaluated models, while \ref{ssec:operationsOnLatent} explains the general concept of how the latent encoding is interpreted to generate the representation of the relit image. It continues with sections providing details about variations of the architecture --~different operations related to the latent representation as well as objective functions used. General schemes of the architectures are enclosed in appendix (see figures \ref{fig:schem1}, \ref{fig:schem2} and \ref{fig:schem3}). Lastly in this section, \ref{ssec:evaluation} describes metrics for model evaluation. The described network components have been implemented using PyTorch \cite{pytorch}.

\subsection{Dataset}
\label{ssec:dataset}

%\subsubsection{Dataset description} %Can we have those titles, maybe as \paragraph{Dataset description} or \subparagraph{Dataset description}? ~Martin
The VIDIT dataset \cite{vidit} used is split into train and validation datasets. The former contains \num{12120} images rendered in 9 different artificial locations, the latter contains \num{1880} images in 3 locations. In each location there are multiple scenes (i.e. locations and orientations of the camera from which the images where rendered). Each one is illuminated under 40 different lighting conditions: 8 cardinal and intercardinal geographic directions from which the light source is illuminating the scene, relative to the camera, and 5 color temperatures of the illuminant, varying from 2500\si{\kelvin} to 6500\si{\kelvin} in 1000\si{\kelvin} increments.

%\subsubsection{Notations} %Same ~Martin
The deep learning network we used to solve the image relighting task takes two images as input: the original image $I$ in which lighting conditions should be changed and the target image $T$ from which the lighting conditions should be acquired. For a given input scene $S_I$ and input illumination $L_I$, we load the corresponding input image $I$. We proceed similarly for the target image $T$ rendered in scene $S_T$ and with target illumination $L_T$. The ground-truth image $G(I,T)$ corresponds to the input scene $S_{G(I,T)} = S_I$ and the target illumination $L_{G(I,T)} = L_T$. Thanks to the construction of the dataset, the ground-truth image $G(I,T)$ always exists. All images are resized from $1024 \times 1024$ to $256 \times 256$ pixels.
The illumination space is the Cartesian product $\left\{L_I\right\}_{\text{image } I} = C \times D$ of the light color temperature set $C = \{2500\si{\kelvin}, 3500\si{\kelvin}, 4500\si{\kelvin}, 5500\si{\kelvin}, 6500\si{\kelvin}\}$ and the light direction set $D = \{NW, N, NE, E, SE, S, SW, W\}$. Therefore, for any image $I$, $L_I = \left(c_I, d_I\right)$ where $c_I \in C$ is the light color temperature in image $I$ and $d_I \in D$ the light direction.

% TODO is that true for all the experiments (pairing strategies?) ~Kuba
From the dataset we are able to construct almost 147 million train dataset pairs and about \num{3.5} million validation dataset pairs. Such a big dataset is intractable for our training capacities; therefore, we firstly limit the dataset by restricting paired images to come from different scenes and to be rendered under different light directions. This new dataset is still very big, and we used two strategies during our training experiments: either we further subsample randomly the created set of image pairs to obtain a smaller training dataset (e.g. 300 thousand image pairs) or we use the entire dataset and thus train without the notion of multiple epochs.

\subsection{Encoder/decoder schema} % I would prefer Encoder/decoder schema over Autoencoder schema: Autoencoder may implicitly suggest groundtruth=input and latent size (chan x width x height) smaller than input size ~Martin
\label{ssec:autoencoder}
\begin{figure}[ht]
    \centering
    \includegraphics[width=1\textwidth]{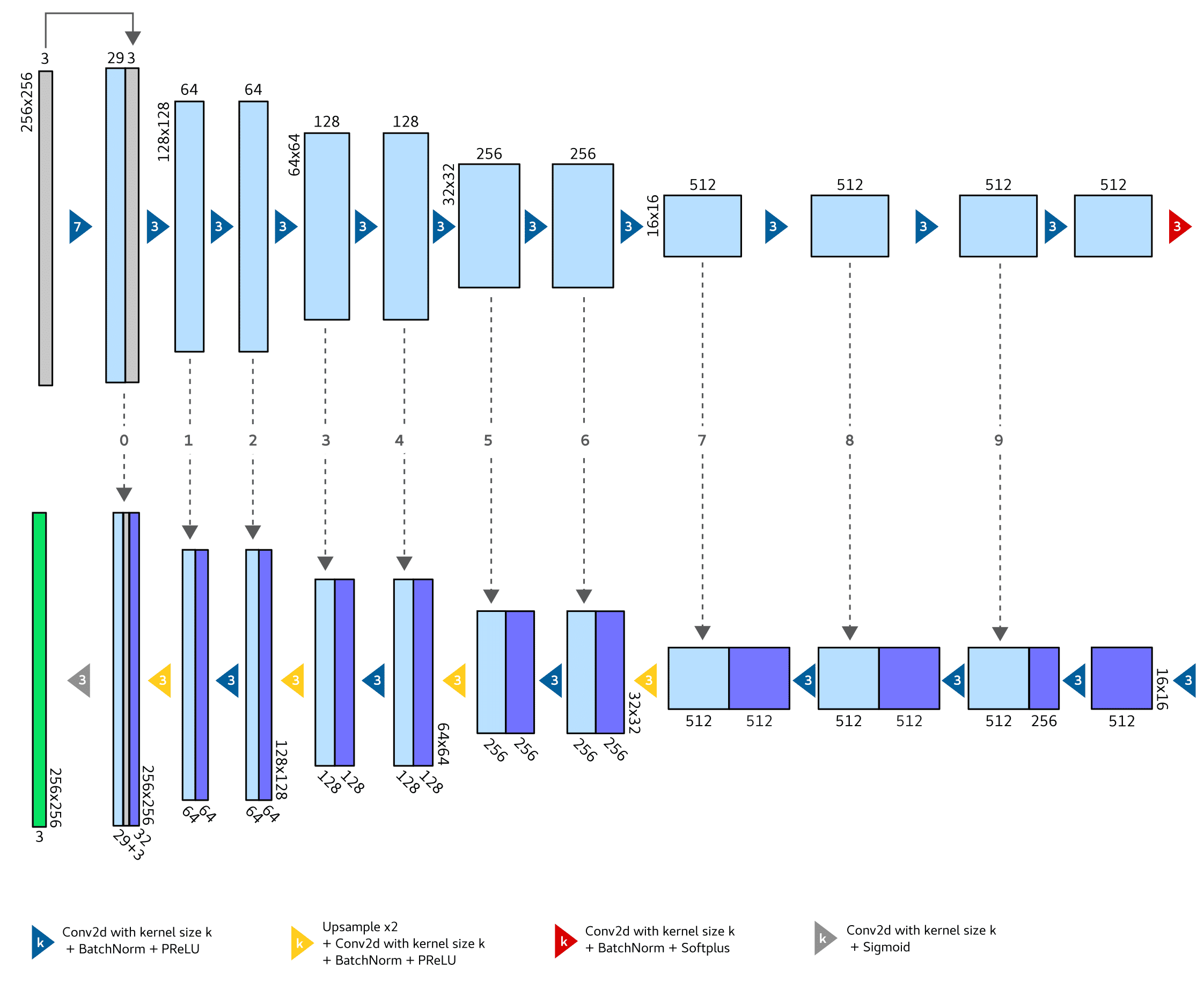}
    \caption{Architecture of the encoder (top) and the decoder (bottom) from \cite{portrait} shared between architectures of our models. Rectangular boxes indicate activation layers with given number of channels (written over the top or below the bottom of the box) and of specified spatial dimensions (written vertically on the side). Gray dashed arrows indicate the skip connections between the encoder and the decoder. The illustration is based on the one in \cite{portrait} with a slightly different graphical style.}
    \label{fig:architectureEncoderDecoder}
\end{figure}
The architectures of our models use the design presented in \cite{portrait} -- a U-Net-like autoencoder with 11 layers for both the encoder and decoder parts -- as depicted in the figure \ref{fig:architectureEncoderDecoder}. In our networks we use batch normalization instead of group normalization.

% \subsubsection{Checkerboard artifacts}
\begin{figure}[hb]
    \centering
    \includegraphics[width=1\textwidth]{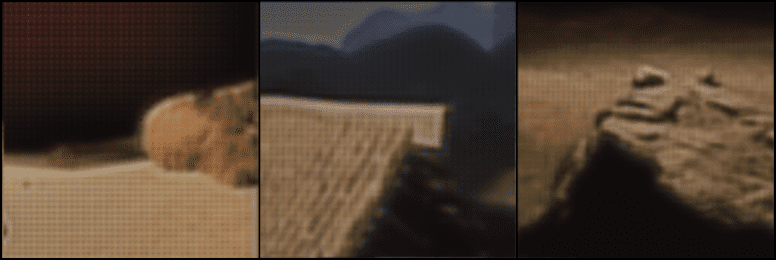}
    \caption{``Checkerboard'' artefacts appearing in the relit images. Images are zoomed-in fragments of the original ones.}
    \label{fig:checkerboard}
\end{figure}
Furthermore, during our experiments we have identified the emergence of ``checkerboard'' artefacts in generated images -- as depicted in the figure \ref{fig:checkerboard}. To solve that issue, the transposed convolution layers have been replaced by successive upsampling and convolution, as suggested in \cite{dlnotes}.
% \subsubsection{...}
The described encoder and decoder parts were shared among all the models presented in this report.

% \subsubsection{...}
In our task the network's input consists of two images $I$ and $T$ and from them the $\hat{G}(I, T)$ image is constructed that aims to approximate actual $G(I, T)$. It differs in that sense from the original problem of \cite{portrait}, where there is a single input image and the target illumination is represented by a small environment map image. For our application, the target illumination conditions are not explicitly known, but are ``encoded'' in the target image $T$. Therefore, in our networks we use a siamese \cite{siamese} encoder for $I$ and $T$ that transforms two images into their respective latent representations. After appropriate transformations on the encodings, a single representation is decoded into an output image. During the decoding, the skip connections from the encoder are used as in the original paper.

% \subssubection{Main loss function} losses will be described within relative architectures I think ~Kuba
% We aim to have $\hat{G} \approx G$ (ie. for any $I$ and $T$, $\hat{G}(I,T) \approx G(I,T)$). Therefore, as a main objective loss, we choose the $\ell_1$ or $\ell_2$ error between the relit image $\hat{G}(I,T)$ produced by the network and the ground-truth image $G(I,T)$. We call it the \texttt{ReconstructionLoss} and we have $\texttt{ReconstructionLoss}_{\ell_2}(G(I,T), \hat{G}(I,T)) = \mathit{\ell_2}(G(I,T), \hat{G}(I,T))$ for instance.

\subsection{Operations on the latent space representation}
\label{ssec:operationsOnLatent}
In the portrait relighting paper \cite{portrait} the latent representation is exchanged for a target environment map that represents the illumination conditions under which the person on the input image should be portrayed. In our models, we either encode in the latent representation only the light conditions of the given image or both the information about the scene content and the illumination. 

From the two encoded images $I$ and $T$, only one encoding representing the original image under the new light conditions should be fed to the decoder. In the scenario in which we treat the latent encoding as representing solely illumination conditions for the given image, the decoder will be provided with the encoding of $T$ in conjunction with skip connections from the encoder of $I$. Otherwise, we split the latent representation into two parts -- scene encoding and light conditions encoding. In all our experiments, we try to guide illumination information encoding only, either by forcing them to take a specific form of artificial generated environment maps (\ref{ssec:groundtruthEnvmap}) or by predicting from it directly light condition properties (\ref{ssec:illuminationPredicter}). In all these variations, the encoding fed to the decoder consists of the scene latent representation (if present) from $I$ and light latent representation encoded from $T$. This way, we aim to keep the content of the original scene (the objects portrayed in the scene) but to present them under illumination conditions from the target image. The information about the objects portrayed in the scene also comes through the skip connections marked in the figure \ref{fig:architectureEncoderDecoder}.

A justification for this approach can be potentially found in the findings of the authors of \cite{semantic}. In their experiments with GAN networks, they show that the learned layers can be grouped in a way in which the activations (intermittent encodings) represent particular hierarchically organized semantic properties of the generated scenes. Most importantly, it is shown that deep layers are responsible for the ``color scheme'' of the generated image, which is of particular interest in our use case. That being said we can expect that the light encoding in the latent representation will correspond to image features that we want to transfer from the target image. At the same time, the skip connections should provide the decoder with intermittent encodings representing geometrical properties of the portrayed scene, such as objects present in the image. We also experiment with approaches that do not enforce any representation on the latent space, and expect it will contain some information about the scene content.

\subsection{Output and latent representation objective functions}

\subsubsection{Abandoned initial concept of light/scene latent loss}
\label{ssec:latentloss}
Our first attempt consisted in having a light latent loss $\mathcal{L}_l$ and a scene latent loss $\mathcal{L}_s$. 

The idea is that two images $I_1$, $I_2$ with the same scene $S_{I_1}=S_{I_2}$ should have similar scene latent variables $\widetilde{\Phi}^{S}_{I_1} \approx \widetilde{\Phi}^{S}_{I_2}$ and two images $I_1$, $I_3$ with different scenes $S_{I_1} \neq S_{I_3}$ should have different scene latent variables: for some distances $\mathit{distance_1}$ and $\mathit{distance_2}$, the intuition is that $\mathit{distance_1}(\widetilde{\Phi}^{S}_{I_1}, \widetilde{\Phi}^{S}_{I_2})$ should somewhat increase as $\mathit{distance_2}\left(S_{I_1}, S_{I_2}\right)$ increases.

Similarly, two images $I_1$, $I_2$ with the same illumination $L_{I_1}=L_{I_2}$ should have similar light latent variables $\widetilde{\Phi}^{L}_{I_1} \approx \widetilde{\Phi}^{L}_{I_2}$: for some distances $\mathit{distance_3}$ and $\mathit{distance_4}$, the intuition is that $\mathit{distance_3}(\widetilde{\Phi}^{L}_{I_1}, \widetilde{\Phi}^{L}_{I_2})$ should somewhat increase as $\mathit{distance_4}\left(L_{I_1}, L_{I_2}\right)$ increases.

The input $I$, the ground-truth $G\left(I,T\right)$, and the relit $\hat G\left(I,T\right)$ images should also have similar scene latent variables $\widetilde{\Phi}^{S}_{I} \approx \widetilde{\Phi}^{S}_{G\left(I,T\right)} \approx \widetilde{\Phi}^{S}_{\hat G\left(I,T\right)}$ since $S_I = S_{G\left(I,T\right)}$. Likewise, we have $L_T = L_{G\left(I,T\right)}$, so we should have $\widetilde{\Phi}^{L}_{T} \approx \widetilde{\Phi}^{L}_{G\left(I,T\right)} \approx \widetilde{\Phi}^{L}_{\hat G\left(I,T\right)}$.

We tried to define a $\texttt{SceneLatentLoss}$ and a $\texttt{LightLatentLoss}$:
\begin{align*}
    \mathcal{L}_{s}^{(1)}\left(I, G\left(I,T\right), \hat G\left(I,T\right)\right) 
    &= \frac{1}{3} \times \ell_2 \left(\widetilde{\Phi}^{S}_{I}, \widetilde{\Phi}^{S}_{G\left(I,T\right)}\right) \\
    &+ \frac{1}{3} \times \ell_2 \left(\widetilde{\Phi}^{S}_{I}, \widetilde{\Phi}^{S}_{\hat G\left(I,T\right)}\right) \\
    &+ \frac{1}{3} \times \ell_2 \left(\widetilde{\Phi}^{S}_{\hat G\left(I,T\right)}, \widetilde{\Phi}^{S}_{G\left(I,T\right)}\right) \\
    \mathcal{L}_{l}^{(1)}\left(T, G\left(I,T\right), \hat G\left(I,T\right)\right) 
    &= \frac{1}{3} \times \ell_2 \left(\widetilde{\Phi}^{L}_{T}, \widetilde{\Phi}^{L}_{G\left(I,T\right)}\right) \\
    &+ \frac{1}{3} \times \ell_2 \left(\widetilde{\Phi}^{L}_{T}, \widetilde{\Phi}^{L}_{\hat G\left(I,T\right)}\right) \\
    &+ \frac{1}{3} \times \ell_2 \left(\widetilde{\Phi}^{L}_{\hat G\left(I,T\right)}, \widetilde{\Phi}^{L}_{G\left(I,T\right)}\right)
\end{align*}
However, with such losses, the network just learns to have a very small variation in $I \mapsto \widetilde{\Phi}^{S}_{I}$ and images with different scenes (respectively illumination) do not really have different scene latent variables (respectively illumination latent variables). In order to avoid this issue, we defined:
\begin{align*}
    \mathcal{L}_{s}^{(2)}\left(I, T, G\left(I,T\right), \hat G\left(I,T\right)\right) 
    &= \frac{\mathcal{L}_{s}^{(1)}\left(I, G\left(I,T\right), \hat G\left(I,T\right)\right)}{\ell_2 \left(\widetilde{\Phi}^{S}_{T}, \widetilde{\Phi}^{S}_{ G\left(I,T\right)}\right)}  \\
    \mathcal{L}_{l}^{(2)}\left(I, T, G\left(I,T\right), \hat G\left(I,T\right)\right) 
    &= \frac{\mathcal{L}_{l}^{(1)}\left(T, G\left(I,T\right), \hat G\left(I,T\right)\right)}{\ell_2 \left(\widetilde{\Phi}^{L}_{I}, \widetilde{\Phi}^{L}_{ G\left(I,T\right)}\right)} 
\end{align*}
as we do not have the constraints $S_T = S_{G\left(I,T\right)}$ nor $L_I = L_{G\left(I,T\right)}$. $\ell_2 \left(\widetilde{\Phi}^{S}_{T}, \widetilde{\Phi}^{S}_{ G\left(I,T\right)}\right)$ and $\ell_2\left(\widetilde{\Phi}^{L}_{I}, \widetilde{\Phi}^{L}_{ G\left(I,T\right)}\right)$ should be interpreted as reference to avoid $I \mapsto \widetilde{\Phi}^{S}_{I}$ to have a very low variation.

This does not work well, and we instead used $\mathcal{L}_{s}^{(1)}$ and $\mathcal{L}_{l}^{(1)}$ as metrics (see \ref{list:metrics}) and did not use them as objective functions.

\subsubsection{Illumination predicter}
\label{ssec:illuminationPredicter}
This method aims to ``force'' the relit image generator to put useful information about the illumination in the light latent variable. It consists of a small fully-connected network. The number of input units is the size of the light latent space. It contains two hidden layers of 20 and 10 units, and two output units. The two output units represent the predicted light direction $\hat{d_I}$ and the predicted light color temperature $\hat{c_I}$.

As we wish to get a predicted light color temperature $\hat{c_I}$ close to the true color temperature $c_I$, we introduce a new loss component: $\mathcal{L}_{c}(c_I, \hat{c_I})$ is the $\ell_2$-error between the true light color temperature $c_I$ and the predicted light color temperature $\hat{c_I}$. It is defined as follows: 

$$\mathcal{L}_{c}(c_I, \hat{c_I}) = \frac{1}{2000^2} \times \ell_2 (c_I-\hat{c_I})$$

Similarly, we want the predicted light direction $\hat{d_I}$ to be close to the true light direction $d_I$, and we introduce $\mathcal{L}_{d}(d_I, \hat{d_I}) = \mathcal{L}_{cos}(d_I, \hat{d_I})$, a cosine-based angular error between the true light direction $d_I$ and the predicted light direction $\hat{d_I}$. The cosine loss is defined as follows: 

\begin{equation}
\label{eq:cosLoss}
    \mathcal{L}_{cos}(x, \hat{x}) = 1 - \cos\left(\pi \times \frac{x-\hat{x}}{\ang{180}}\right)
\end{equation}

\noindent Note that $\mathcal{L}_{d} = 1$ may indicate that the light direction cannot be inferred from the light latent variable as it is the expected value for this loss for random prediction of light direction:

$$\mathbbm{E}_{\left(\hat{d_I}~\Mod{~360}\right) \sim \mathcal{U}(\ang{0}, \ang{360})}{ \mathcal{L}_{d}\left(d_I, \hat{d_I}\right) } = 1$$

\subsubsection{Generated environment map: a ground-truth for light latent variable}
\label{ssec:groundtruthEnvmap}
One of the key ideas presented in \cite{portrait} is their interpretation of the light latent representation and consequently the way in which it is used as one of the loss components. Their aim is to model the ``environment map'' of the image, which is a small $16 \times 32$ pixel image, through the latent representation. To create the ``ground-truth'' environment maps that will be learnt by the network in the latent representation, the environment maps, used for creation of their dataset, are ``mapped back to the latitude-longitude format'' of given small image size by modelling LED lights (used to illuminate people in portrait photographs) as Gaussian distributions with the means indicating their positions \cite{portrait}.

Another important processing step presented in the mentioned paper is how to go from a multi-channel tensor in the latent space to a three-channel, $16 \times 32$ RGB image that will be compared with the ground-truth environment map. To achieve that goal, they use a weighted pooling layer, a concept introduced in \cite{constancy}, cited by the authors of \cite{portrait}, about estimating color constancy of the image. Given a latent variable tensor $\Phi \in \mathbb{R}^{2028 \times 16 \times 16}$ (first dimension is the number of channels, two last ones are spatial dimension) it is split into four tensors $\mathbb{R}^{512 \times 16 \times 16}$: $\Phi_c$, $\Phi_R$, $\Phi_G$, $\Phi_B$. The first of them represents pooling weights that will be applied to each of the other tensors representing respectively the contents of the red, green and blue channels of the estimated color map. Lastly, the weighting pooling is applied by an element-wise multiplication of $\Phi_c$ with each of the tensors representing color channels and summing across spatial dimensions so that three $\mathbb{R}^{512 \times 1 \times 1}$ tensors are obtained: $\hat{\Phi}_R$, $\hat{\Phi}_G$, $\hat{\Phi}_B$.

The estimated environment map channels content can be rearranged, so that each of them forms a $16 \times 32$ pixel image channel of the estimated environment map $\hat{\Phi}^L_I$ of the image $I$. Weighted pooling can be interpreted as follows: to obtain the value of each pixel in the estimated environment map, all pixels within one channel vote, with learned weights, for the final value. Since those pixels are representing some spatial ``patches'' \cite{constancy} in the original image due to the nature of convolutional neural networks, the weights should represent the learned importance of certain image patches in estimating light condition information \cite{portrait,constancy}.

We have used those concepts in our architecture as well, although in a simplified manner. We have created two models: one which follows exactly the latent representation interpretation and processing from the original paper \cite{portrait} and second that assigns part of the obtained latent variable to represent environment map in the given image, while the rest of the encoding is not guided by any loss and aims to convey information about content of the image other than illumination, such as the geometry of the objects. In both cases weighted pooling is used for entire or significant part of the light latent representation.

Training with these models requires not only the ground-truth for the relit image, but also a desired light latent space representation -- the environment maps. Since our dataset generation is inherently different from the one presented in \cite{portrait} we decided to generate our ground-truth environment maps based on the known light properties: its color temperature and direction. Color temperature has been translated into RGB values with the use of the Colour-Science \cite{colourscience} Python library. We represent light direction as a Gaussian distribution over value/brightness component of the light color in HSL color space, with its mean $\mu$ localized on the horizontal axis of the generated image and standard deviation $\sigma$ equal to 10\% of the environment map height (see figures \ref{fig:envmaps} and \ref{fig:envmapsScene} in the appendix for an example).

\subsubsection{Latent representation as environment map estimation only}

In this scenario following the original paper, the entire latent representation is used for light conditions representation. This is done by teaching the network to learn how to estimate the environment map by adding the loss between the pair of ground-truth generated environment map and one produced by the model in the latent space. We use the same loss as was used in \cite{portrait} for that goal:

\begin{equation}
\label{eq:envmapLoss}
    \mathcal{L}_{e}({\Phi}^{L}_{I}, \widetilde{{\Phi}}^{L}_{I}) = \left\Vert \log\left(1 + {\Phi}^{L}_{I}\right) - \log\left(1 + \widetilde{{\Phi}}^{L}_{I}\right) \right\Vert^{2}_{2}
\end{equation}

\noindent The second loss between the generated relit image and ground-truth image, the $\ell_1$-reconstruction loss (naming after e.g. \cite{underexposed}), is also used same as in \cite{portrait} except that we do not mask the image to a restricted part that will be considered for the optimization:

\begin{equation}
\label{eq:reconstructionLoss}
    \mathcal{L}_{r}(\hat G(I, T), G(I, T)) = \left\Vert \hat G(I,T) - G(I,T) \right\Vert_{1}
\end{equation}

\subsubsection{Latent representation as scene and light representation}

In this variant, only a part of the latent variable is used to estimate the environment map for the image, while the remaining part is not guided by any objective function. We expect the non-guided part to form the encoding of the scene content, while the guided environment map estimation to represent the light conditions. In such an encoding, the latent tensor $\Phi \in \mathbb{R}^{2052 \times 16 \times 16}$ is split into scene representation $\Phi^{S} \in \mathbb{R}^{1024 \times 16 \times 16}$ and light conditions representation $\Phi^{L} \in \mathbb{R}^{1028 \times 16 \times 16}$. In the latter, half of the channels are used for weighted pooling weights. The other half is used for light color estimation in HSL color space: one channel for hue estimation, one for saturation estimation and remaining 512 channels (corresponding to $16 \times 32$ image of the estimated environment map) for the value/brightness estimation. In this variant, ground-truth environment map is an $\mathbb{R}^{514 \times 1 \times 1}$ tensor, with first and second element containing the hue and saturation values respectively and the remaining 512 elements containing the brightness values. To establish a loss between estimated and ground-truth environment map we use three loss components: cosine distance for hue estimation (\ref{eq:cosLoss}), mean-squared error for saturation estimation and loss from equation (\ref{eq:envmapLoss}) for brightness estimation. In this variant, instead of $\ell_1$- or $\ell_2$-reconstruction loss, the relit image is compared with the ground-truth image through LPIPS \cite{lpips} measure which is minimized during the training.

\subsection{Model evaluation}
\label{ssec:evaluation}
The scenes of the dataset have been split into a train scenes set $\Sigma_\mathit{train}$ and an evaluation scenes set $\Sigma_\mathit{eval}$. We have $\Sigma = \Sigma_\mathit{train} \uplus \Sigma_\mathit{eval}$. For our methods we call the train dataset $\mathit{Train}$ the set of triplets $(I, T, G(I, T))$ such that the scenes come from the train scenes set $(S_I, S_T) \in (\Sigma_\mathit{train})^2$:

$$\mathit{Train} = \{(I, T, G(I, T)) | (S_I, S_T) \in (\Sigma_\mathit{train})^2\}.$$

\noindent Similarly, the evaluation set is:
$$\mathit{Eval} = \{(I, T, G(I, T)) | (S_I, S_T) \in (\Sigma_\mathit{eval})^2\}.$$

In order to assess the strengths and weak points of one model (one presented in \ref{sssec:illuminationPredicterModel}), we have created some evaluation subsets:

% https://tex.stackexchange.com/a/241081
\begin{equation*}
\begin{aligned}
    \mathit{Eval}_{c_I = c_T} ={} & \{(I, T, G(I, T)) | (S_I, S_T) \in (\Sigma_{eval})^2 \text{ and } c_I = c_T\}, \\
    \mathit{Eval}_{d_I = d_T} ={} & \{(I, T, G(I, T)) | (S_I, S_T) \in (\Sigma_{eval})^2 \text{ and } d_I = d_T\}.
\end{aligned}    
\end{equation*}

We now introduce some metrics we used:
\begin{itemize}
\label{list:metrics}
    \item We measure the distances between the scene latent variables: $$\ell_2 \left(\widetilde{\Phi}^{S}_{I_1}, \widetilde{\Phi}^{S}_{I_2}\right) \text{ for } I_1, I_2 \in \left\{I, T, G(I,T), \hat G(I,T)\right\}$$
    This allows us to check -- as explained in \ref{ssec:latentloss} -- if images with similar scenes have similar scene latent variables.
    \item Similarly, we measure the distances between the illumination latent variables: $$\ell_2 \left(\widetilde{\Phi}^{L}_{I_1}, \widetilde{\Phi}^{L}_{I_2}\right) \text{ for } I_1, I_2 \in \left\{I, T, G(I,T), \hat G(I,T)\right\} $$
    This allows us to check -- as explained in \ref{ssec:latentloss} -- if images with similar illuminations have similar light latent variables.
    \item SSIM: it measures the structural similarity \cite{ssim} between 2 images. We aim to have high SSIM values (high similarity) between the ground-truth $G(I,T)$ and the relit $\hat G(I,T)$ images.
    \item PSNR: the peak signal-to-noise ratio is also a traditional metric for images. However, it is known to be a weaker metric \cite{psnrvsssim}.
    \item LPIPS: a perceptual metric that aims to be close to human perception of similarity. It stands for Learned Perceptual Image Patch Similarity and was first described in \cite{lpips}. We aim to have low LPIPS values (high similarity) between the ground-truth $G(I,T)$ and the relit $\hat G(I,T)$ images.
    \item \texttt{Score}: we build a \texttt{Score} metric to assess whether the network is doing better than the identity mapping $\hat G(I,T) = I$. Therefore, we divide the value of the $\ell_2(G(I,T), \hat{G}(I,T))$ by the value that would have been the $\ell_2$-reconstruction loss if we had $\hat G(I,T) = I$. Therefore, we define $$\texttt{Score}_{\ell_2}(I, G(I,T), \hat G(I,T)) = \frac{\mathit{\ell_2}(G(I,T), \hat{G}(I,T))}{\mathit{\ell_2}(G(I,T),I)}$$
\end{itemize}

% I remove it as we didn't do that much
% In order to reject some models that are too weak, we first see if our model is able to over-fit on 1-to-1 transfer. Then we see if it can overfit on N-to-N transfer with 1 location. The we finally train our model on N-to-N transfer with all locations.

In order to follow the evolution of our models as it progressed, we have used Tensorboard \cite{tensorflow}. Not only we can plot the different metrics, but we can also show some sample images at each iteration. This is done for both train and evaluation sets.

\subsubsection{Make the relit image realistic with GAN}

As a way to make our relit image look more realistic, we used a Generative Adversarial Net (GAN) \cite{gan}. Alongside the generative model $G$, this technique uses a discriminative model $D$ that estimates the probability that a sample is created rather than authentic. The discriminator $D$ learns a mapping $s \rightarrow [0,1]$ where 1 means the input $s$ is real, and 0 means it was created by $G$. Those two models are jointly trained to help $G$ produce convincing and realistic samples.

The paper Image-to-Image Translation with Conditional Adversarial Networks \cite{patchgan} uses a conditional GAN with an interesting discriminator architecture they named ``PatchGAN''. Instead of looking at the whole image to determine if it is real or fake, the discriminator $D$ classify each $N \times N$ patch in an image, and then averages all the results to produce its final output. Thus, this discriminator only corrects high-frequency structures, while the low-frequency correctness is handled by an $\ell_1$-loss. This discriminator allows them to produce images that are less blurry than images produced with a regular GAN, and the paper shows that it can be used to produce reasonable results on a wide variety of tasks. 
% TO DO : répétition litterature review. Remove/skim/it's fine ? remove if it's repeated ~Kuba

%\subsubsection{Use GAN to remove reconstruction loss}
%TO DO : delete ?

\section{Results}

As we were making choices regarding our models based on the observations of how well they performed on the evaluation set, all the results displayed in this section -- computed with the evaluation set -- should be confirmed using a test set we neither used for training nor for model selection and hyper-parameter tuning.

\subsection{Results related to light latent variable}

\subsubsection{Illumination predicter: light color temperature}
\label{sssec:illuminationPredicterModel}

\begin{figure}[H]
    \centering
    \begin{minipage}[c]{0.49\linewidth}
        \includegraphics[width=1\textwidth]{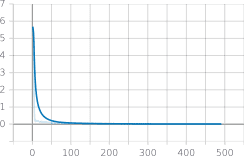}
    \end{minipage} \hfill
    \begin{minipage}[c]{0.49\linewidth}
        \includegraphics[width=1\textwidth]{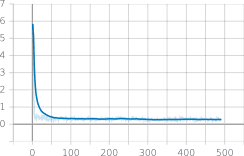}
    \end{minipage}
    \begin{minipage}[c]{0.49\linewidth}
        \includegraphics[width=1\textwidth]{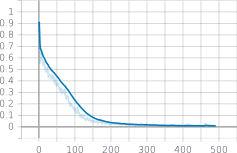}
    \end{minipage} \hfill
    \begin{minipage}[c]{0.49\linewidth}
        \includegraphics[width=1\textwidth]{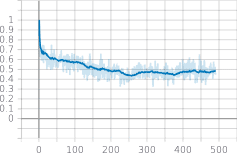}
    \end{minipage}
    \caption{First row: evolution of the light color temperature prediction loss $\mathcal{L}_{c}$ on the train set $\mathit{Train}$ (left) and on the evaluation set $\mathit{Eval}$ (right). Second row: evolution of the light direction prediction loss $\mathcal{L}_{d}$ on the train set (left) and on the evaluation set (right).}
    \label{fig:predictionLoss}
\end{figure}

In the model with illumination predicter where the latent variable is split into a light and a scene latent variables, we observe the evolution of the light color temperature prediction loss $\mathcal{L}_{c}$ depicted in the first row of the figure \ref{fig:predictionLoss}. In the train set, the (smoothed) value of $\mathcal{L}_{c}$ goes down to 0.01 and in the different evaluations sets, the (smoothed) value goes down to 0.26. This means the error of light color temperature prediction is roughly $RMSE \approx \sqrt{ 2000 \times 0.26 } \approx 23 \si{\kelvin}$. Therefore, the encoder part of the relit images generator successfully learned to put the color temperature information in the light latent variable.

\subsubsection{Illumination predicter: light direction}

In the same model, we observe the evolution of the light direction prediction loss $\mathcal{L}_{d}$ depicted in the second row of the figure \ref{fig:predictionLoss}. In the train set, the (smoothed) value of $\mathcal{L}_{d}$ goes down to 0.01 and in the different evaluations sets, the (smoothed) value goes down to 0.44. This means the angular error of light direction prediction is roughly $\mathit{arccos}(1-0.44) \approx 0.98 \mathit{~rad} \approx 56^\circ $. Therefore, the encoder part of the relit images generator successfully learned to put some light direction information in the light latent variable, but didn't extract the direction as accurately as the color temperature.

\begin{figure}[H]
    \centering
    \begin{minipage}[c]{0.32\linewidth}
        \includegraphics[width=1\textwidth]{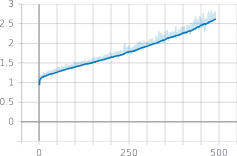}
    \end{minipage} \hfill
    \begin{minipage}[c]{0.32\linewidth}
        \includegraphics[width=1\textwidth]{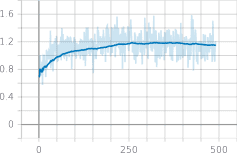}
    \end{minipage}\hfill
    \begin{minipage}[c]{0.32\linewidth}
        \includegraphics[width=1\textwidth]{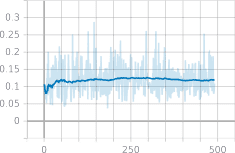}
    \end{minipage}
    \caption{Evolution of the $\texttt{Score}_{\ell_2}$ on the train set $\mathit{Train}$ (left) and on the full evaluation set $\mathit{Eval}$ (middle) and on the evaluation subset $\mathit{Eval}_{d_I = d_T}$ (right).}
    \label{fig:score}
\end{figure}

\begin{figure}[H]
    \centering
    \begin{minipage}[c]{0.22\linewidth}
        \includegraphics[width=1\textwidth]{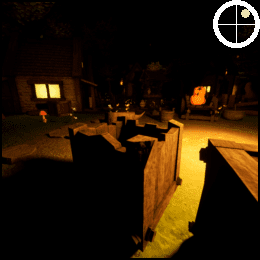}
    \end{minipage} \hfill
    \begin{minipage}[c]{0.22\linewidth}
        \includegraphics[width=1\textwidth]{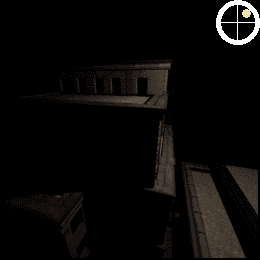}
    \end{minipage}\hfill
    \begin{minipage}[c]{0.22\linewidth}
        \includegraphics[width=1\textwidth]{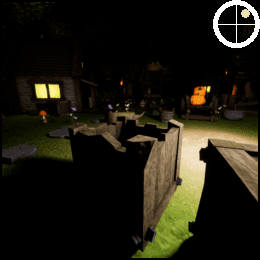}
    \end{minipage}\hfill
    \begin{minipage}[c]{0.22\linewidth}
        \includegraphics[width=1\textwidth]{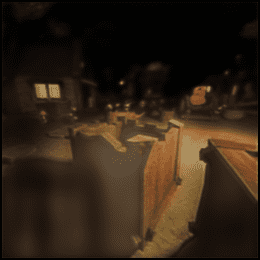}
    \end{minipage}
    \caption{Example of relighting performed on the evaluation subset $\mathit{Eval}_{d_I = d_T}$ (from left to right: input $I$, target $T$ and ground-truth $G(I,T)$ images; relit image $\hat{G}(I,T)$ produced by this model).}
    \label{fig:samedirection}
\end{figure}

With a maximum (smoothed) value for the score of 1.25 on the evaluation set $\mathit{Eval}$ -- as depicted in the figure \ref{fig:score} -- our model does better than the identity mapping $\hat G(I,T) = I$. A main component of those good results comes from the ability to "remove" shadows that the network has learned. We can confidently affirm this thanks to the middling score of 0.14 on the evaluation subset $\mathit{Eval}_{d_I = d_T}$ on which the identity mapping gives really good result and on which removing shadows actually worsen the relit images -- as shown in the figure \ref{fig:samedirection}.


\subsubsection{Environment map estimation models}

\begin{figure}[h]
    \centering
    \begin{minipage}[c]{0.49\linewidth}
        \includegraphics[width=1\textwidth]{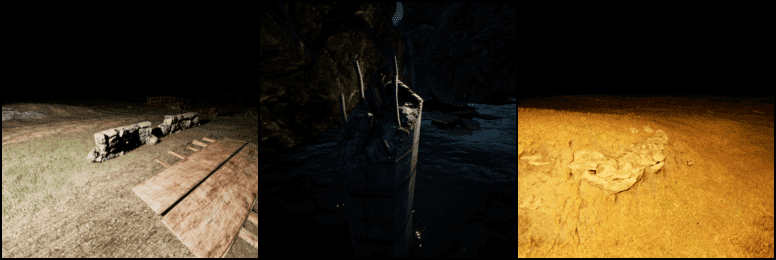}
    \end{minipage} \hfill
    \begin{minipage}[c]{0.49\linewidth}
        \includegraphics[width=1\textwidth]{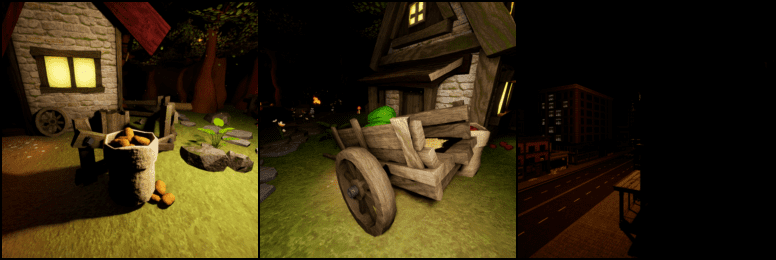}
    \end{minipage}
    \begin{minipage}[c]{0.49\linewidth}
        \includegraphics[width=1\textwidth]{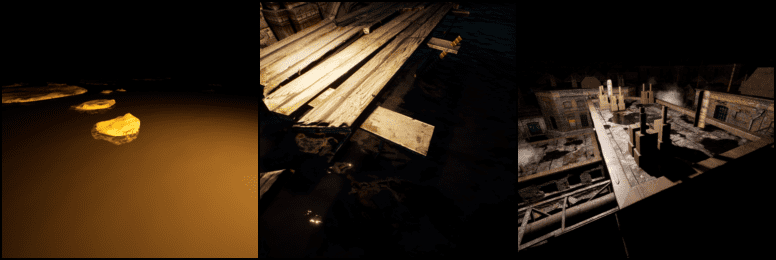}
    \end{minipage} \hfill
    \begin{minipage}[c]{0.49\linewidth}
        \includegraphics[width=1\textwidth]{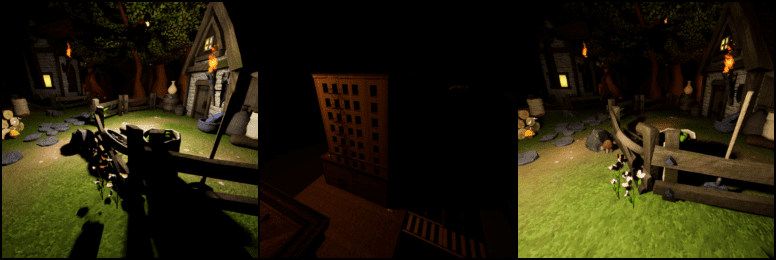}
    \end{minipage}
    \begin{minipage}[c]{0.49\linewidth}
        \includegraphics[width=1\textwidth]{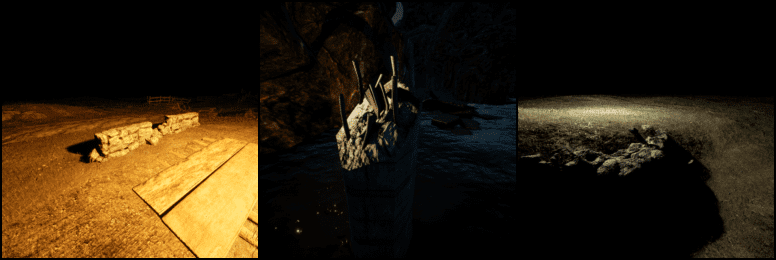}
    \end{minipage} \hfill
    \begin{minipage}[c]{0.49\linewidth}
        \includegraphics[width=1\textwidth]{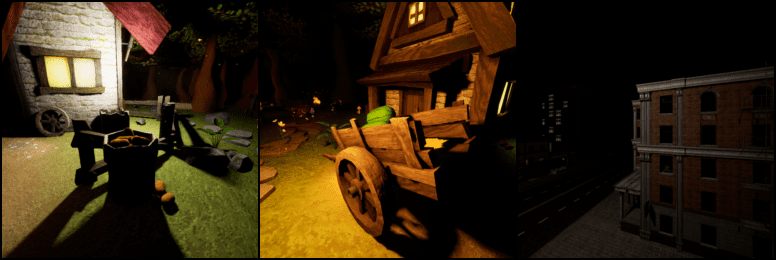}
    \end{minipage}
    \begin{minipage}[c]{0.49\linewidth}
        \includegraphics[width=1\textwidth]{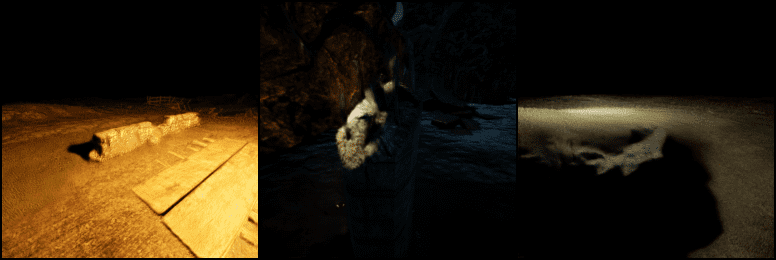}
    \end{minipage} \hfill
    \begin{minipage}[c]{0.49\linewidth}
        \includegraphics[width=1\textwidth]{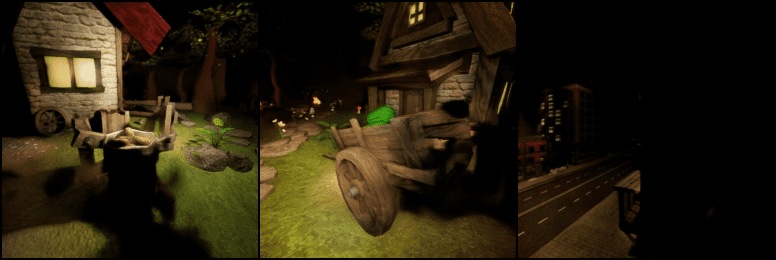}
    \end{minipage}
    \caption{Examples of relighting performed by the model with latent variables representing only the estimated environment map. From top to bottom: input $I$, target $T$, ground-truth $G(I, T)$, relit image $\hat G(I, T)$. Left column contains the samples from the train set in the 9\textsuperscript{th} training epoch, right column presents the samples from the test set produced after 7 epochs of training.}
    \label{fig:resultsEnvmap}
\end{figure}

\textbf{Latent representation as environment map estimation only.} As presented in the figure \ref{fig:resultsEnvmap}, the relighting results are very good for the samples from the training set (left side) but also very poor for the ones from the validation set (right side). This indicates that network seems to have a big enough capacity to perform the relighting, yet it likely over-fits to the training dataset. It might have either over-fitted to the given data samples (in that case it was almost 300 thousand of them\footnote{It might have been almost 200 thousand as well, we have not registered the number of the datasize we used for training.} randomly sub-sampled from the all possible input-target combinations) or to the locations and their content. It should be also noted that for the train samples relighting, the output images loose their details i.e. clear brick edges in the first image or rock texture in the third image.

\begin{figure}[h]
    \centering
    \begin{minipage}[c]{0.49\linewidth}
        \includegraphics[width=1\textwidth]{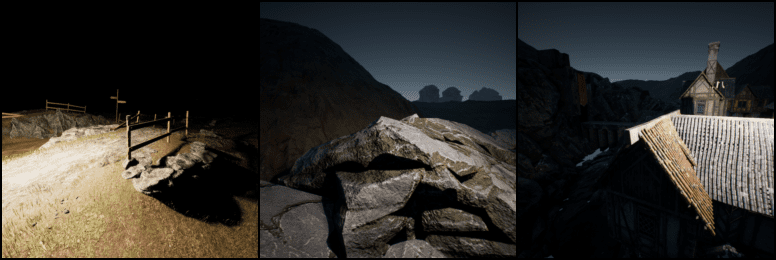}
    \end{minipage} \hfill
    \begin{minipage}[c]{0.49\linewidth}
        \includegraphics[width=1\textwidth]{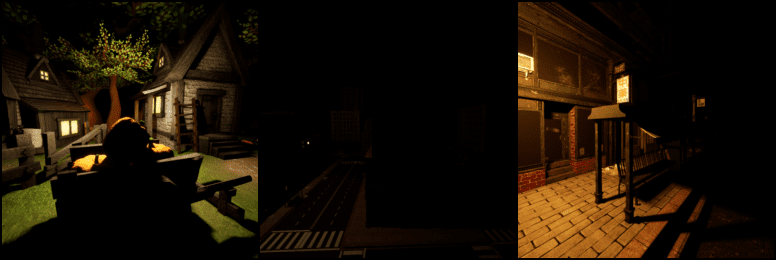}
    \end{minipage}
    \begin{minipage}[c]{0.49\linewidth}
        \includegraphics[width=1\textwidth]{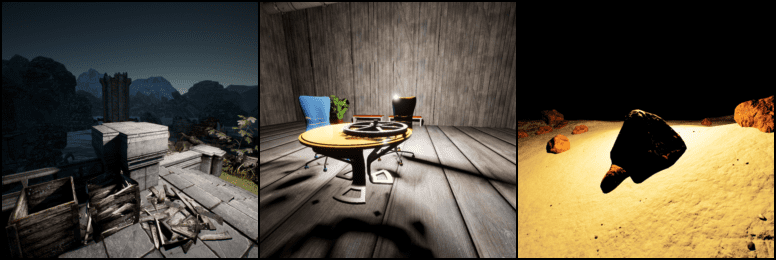}
    \end{minipage} \hfill
    \begin{minipage}[c]{0.49\linewidth}
        \includegraphics[width=1\textwidth]{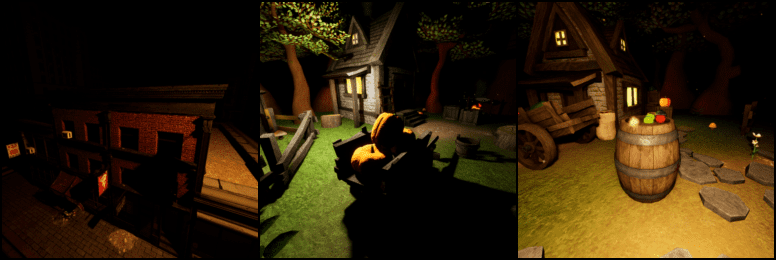}
    \end{minipage}
    \begin{minipage}[c]{0.49\linewidth}
        \includegraphics[width=1\textwidth]{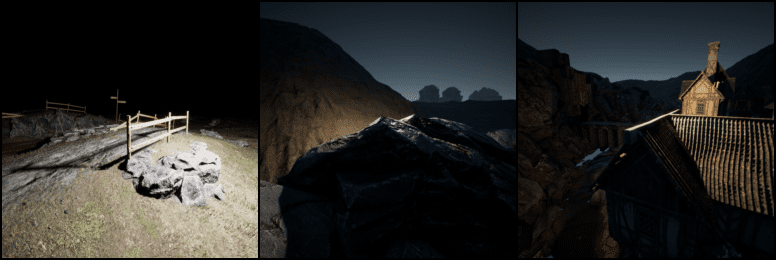}
    \end{minipage} \hfill
    \begin{minipage}[c]{0.49\linewidth}
        \includegraphics[width=1\textwidth]{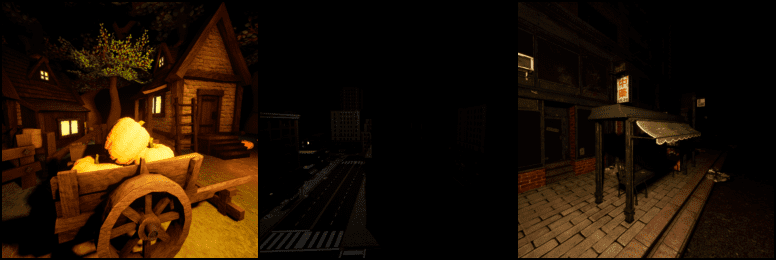}
    \end{minipage}
    \begin{minipage}[c]{0.49\linewidth}
        \includegraphics[width=1\textwidth]{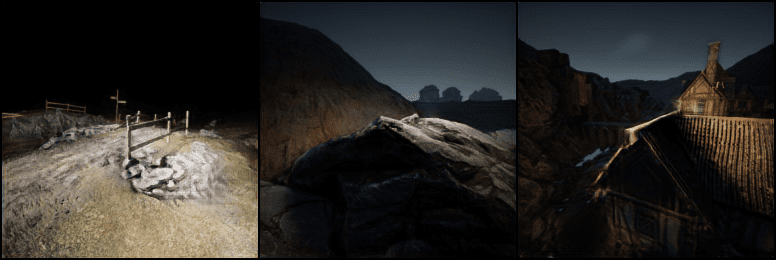}
    \end{minipage} \hfill
    \begin{minipage}[c]{0.49\linewidth}
        \includegraphics[width=1\textwidth]{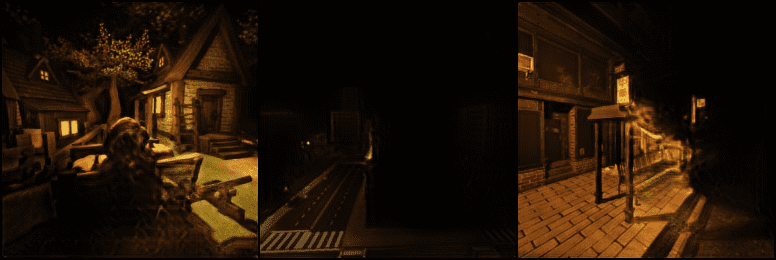}
    \end{minipage}
    \caption{Examples of relighting performed by the model with environment map estimation and both the light and the scene latent representation. From top to bottom: input $I$, target $T$, ground-truth $G(I, T)$, relit image $\hat G(I, T)$. Left column contains the samples from the train set in the 10\textsuperscript{th} training epoch, right column presents the samples from the test set produced after 6 epochs of training.}
    \label{fig:resultsEnvmapScene}
\end{figure}

\textbf{Latent representation as scene and light representation.} The results of the relighting with the model using a light-scene latent split and environment map estimation are presented in the figure \ref{fig:resultsEnvmapScene}. It is important to highlight that for the relighting examples from the training set, the outputs contain much more details and less blurring than the ones from the model described above and presented in the figure \ref{fig:resultsEnvmap}. This can be accounted either to the usage of LPIPS as the reconstruction loss or to the fact that this model dedicates part of the latent variable for the scene encoding. Still, the results for the evaluation set are far from desired. The model manages to put some light (i.e. first image with the car with pumpkins) or to slightly suppress the shadow (i.e. shadow from the post of the stand).

\subsection{Realism of relit images}
Some results obtained using a PatchGAN as a discriminator make us think the use of discriminator might slightly improve our metrics and the realism of the relit images -- as depicted in the appendix, figures \ref{fig:resultsnogan} and \ref{fig:resultsgan}. However, it would be necessary to train 2 models where the only difference is the presence/absence of the discriminator to confirm or disprove that -- by a mistake of using a wrong $\mathcal{L}_{d}$ loss for the training with the discriminator, we failed to compare the two kinds of models.

We also noticed that our model with LPIPS loss and latent representation as scene and light representation produces more realistic images. This can be accounted to both the fact that forcing the latent representation to solely encode the light conditions is not enough to generate good results -- as suggested during the project meetings -- but also to the LPIPS metric that more closely resembles the notion of ``similarity'' than e.g. $\ell_2$-reconstruction loss. In fact, the authors that proposed the LPIPS metric pointed out that the $\ell_2$-error is not very sensitive to blurring, while it is easily noticed by humans \cite{lpips}.

\subsection{Model comparison}

The comparison of the three models presented above is summarized in table \ref{table:results}. The metrics are far from desired higher values for PSNR and SSIM. The models do not differ significantly within the presented scores. The model with the illumination predictor does best with respect to the SSIM, while the highest PSNR value is achieved for the model using latent representation for only environment map estimation, therefore the one that is almost identical to original \cite{portrait}. The lowest (best) LPIPS was achieved by the model with latent split into scene and light representation and environment map estimation. However, it should be noted that this model used LPIPS as its reconstruction loss, therefore this result is not surprising and is not a proper metric to be applied for this particular model.

{\renewcommand{\arraystretch}{1.2}
\begin{table}[h!]
    \begin{center}
         \begin{tabular}{c c c c} 
             \hline
             Metric name & IlluminationPredicter & Envmap & Envmap + scene \\
             \hline\hline
             MSE & 0.0238 & \bfseries 0.0219 & 0.0254 \\
             PSNR & 18.11 \si{dB} & \bfseries 18.66 \si{dB} & 18.10 \si{dB} \\
             \hline
             SSIM & \bfseries 0.3365 & 0.1832 & 0.2988 \\
             \hline
             LPIPS & 0.3268 & 0.2738 & \bfseries 0.2564 \\
             \hline
        \end{tabular}
    \end{center}
    \caption{Quantitative results for the three models presented in the report: \textbf{IlluminationPredicter} (see \ref{ssec:illuminationPredicter}) -- model using a fully-connected network to infer light conditions properties from the latent representation; two models based on environment map estimation in the latent space (see \ref{ssec:groundtruthEnvmap}) -- \textbf{Envmap} for which the encoding represents solely the lighting conditions in the image and \textbf{Envmap + scene} that splits the latent space between the scene and the light representation. The best result for each metric has been bolded. Note that Envmap~+~scene model has been optimized for LPIPS, so its best performance with this metric is not unexpected.}
    \label{table:results}
\end{table}}

\section{Conclusion}

%Good
Our results show qualitative improvement compared to producing just the identity mapping. It is clear that teaching the network to change the illumination color is feasible and much easier than relighting the image with a differently localized light source.

%How
In a very simplified manner, it seems the networks first learn the identity mapping, to put light information in light latent variable and to change color temperature, then to remove shadings, and finally to add shadows (and probably put scene information in scene latent variable). These rough steps seem to have increasing difficulty levels.

%Bad
The main failures of our models are the poor realism of the relit images and the fact they tend to give a bigger edge to shading removal than to real light direction change. The models are sometimes able to remove and add some shading, but it is not consistent nor accurate. We suspect the models are not capable of generalizing to unseen locations -- that would explain the good relighting results obtained on the training dataset, but poor ones on the validation dataset.

%Why
It should be noted that both relit images as targets contain strong, deep shadow areas that are mostly black. This leads to great difficulties and imprecisions during relighting with a modified light direction -- e.g. objects and surfaces in very dark areas should come into light and thus it may be very challenging for the network to ``guess'' what kind of content should be placed there. We can conclude that such networks would need to possess the ability to ``imagine'' appropriate content matching the overall scene. This is certainly a very difficult task, even for humans, which is also ill-posed -- there are potentially many possibilities of filling the dark areas with meaningful and relevant content.

\subsection{Possible improvements}
%Try some variations
It would be interesting to improve our model further by experimenting with more variations over our network. Among others, it would be possible to use PixelShuffle \cite{pixelshuffle} instead of upsampling to address the checkerboard artifacts, to further tune the network hyperparameters (e.g. depth, latent variable size), to use some regularization (as $\ell_2$-regularization), experiment with additional loss factors and functions. It would be also valuable, as suggested, to properly enforce the light-scene latent representation split. In presented architectures the only part of the latent representation that is guided through objective function minimization is the light representation. However, it cannot be assured that the remaining parts of latent representation contain solely information about the scene features nor that they are meaningful. To potentially guarantee that, one possibility would be to enforce scene representation part of the latent variable to be identical for data samples with the same scene under varying light conditions.

%More time to work on GAN
Ideally, we would like the network to learn to produce realistic relit images. Obviously, our results do not seem as such and our method might gain if further experiments with GAN were to be conducted. It would hopefully be able to make relit images look more real, using a relativistic GAN for instance \cite{relativisticgan}. Not only the relit images must be realistic on their own, but they should be consistent with the input and target images as well. While this consistency is ensured by the reconstruction loss, the network tends to remove very dark shadings by replacing them for instance with a more-or-less uniformly colored area, as in the figures \ref{fig:samedirection} and \ref{fig:resultsgan}. It would therefore be interesting to try implementing a conditional GAN \cite{cgan}.

%non deeplearning possibilities
Finally, some very recent similar works, either with non-deep-learning methods -- such as \cite{nodeeplearning} -- or with deep-learning architectures -- such as \cite{daytime} -- gave interesting and quite realistic results. 

\subsection*{Special thanks}
We would like to express special thanks to Majed El Helou, Ruofan Zhou and Sabine S{\"u}sstrunk for their supervision and for their help in this project.

% ---- Bibliography ----
%
% BibTeX users should specify bibliography style 'splncs04'.
% References will then be sorted and formatted in the correct style.
%

% \addbibresource{bibliography.bib}
\bibliographystyle{splncs04}
\bibliography{bibliography.bib}
\clearpage

% Appendix
\appendix

\section{Schemes of architectures for the described models}

Loss denoting in following figures inspired by \cite{portrait}.

\begin{figure}[!ht]
    \centering
    \includegraphics[width=0.7\textwidth]{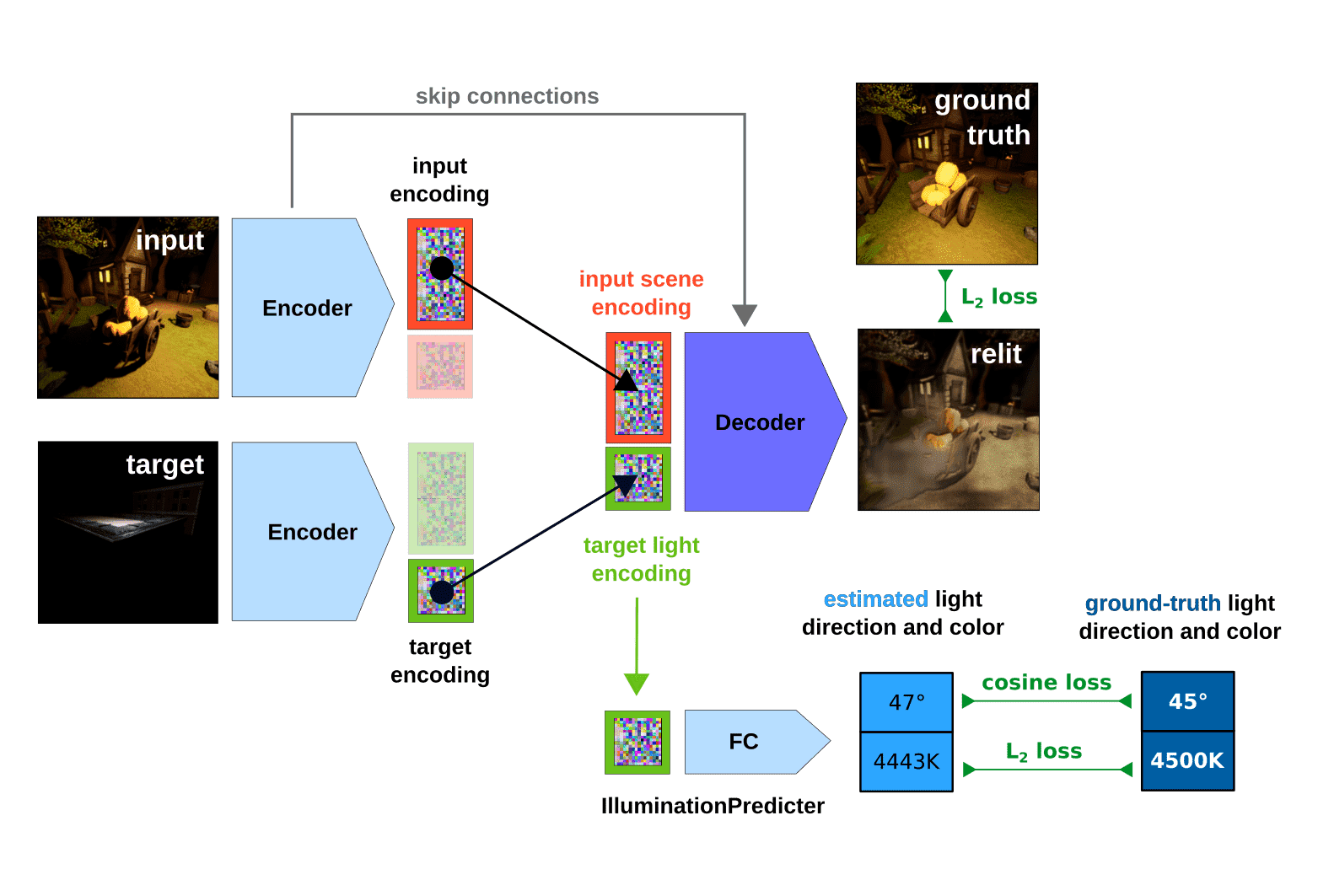}
    \caption{Scheme of the architecture for the model \textbf{IlluminationPredicter} with $\ell_2$-reconstruction loss, latent light-scene split and illumination predicter with light direction and light color temperature predictions auxiliary losses. }
    \label{fig:schem1}
\end{figure}

\begin{figure}[!ht]
    \centering
    \includegraphics[width=0.7\textwidth]{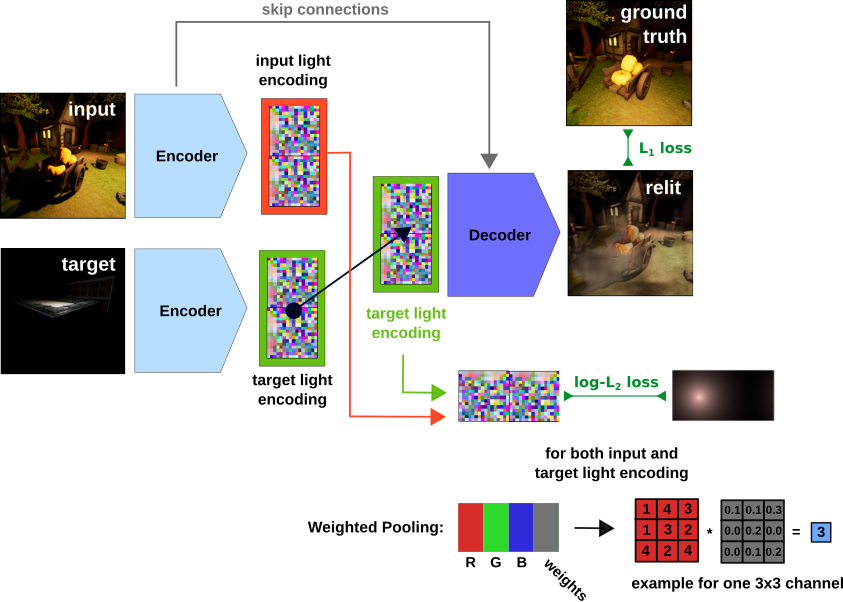}
    \caption{Scheme of the architecture for the model \textbf{Envmap} with $\ell_1$-reconstruction loss, latent light representation only and generated environment maps for $\mathcal{L}_{e}$ auxiliary loss.}
    \label{fig:schem2}
\end{figure}

\begin{figure}[!ht]
    \centering
    \includegraphics[width=0.7\textwidth]{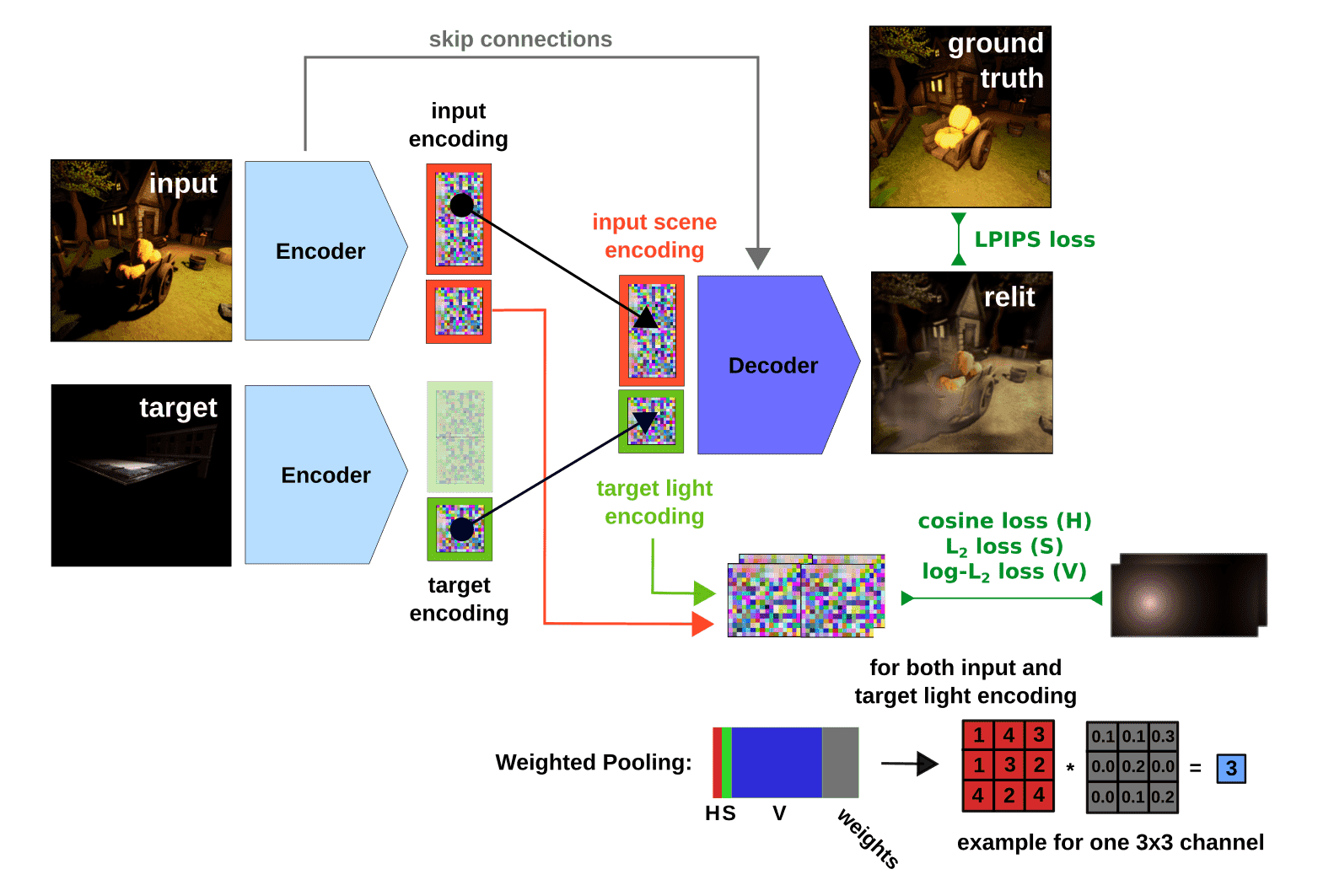}
    \caption{Scheme of the architecture for the model \textbf{Envmap + scene} with LPIPS reconstruction loss, latent light-scene split and combined cosine loss for hue, mean-squared error loss for saturation and $\mathcal{L}_{e}$ loss for brightness of latent variable (environment map).}
    \label{fig:schem3}
\end{figure}

\section{Failed comparison between of a model with and without a discriminator}

\begin{figure}[H]
    \centering
    \begin{minipage}[c]{0.49\linewidth}
        \includegraphics[width=1\textwidth]{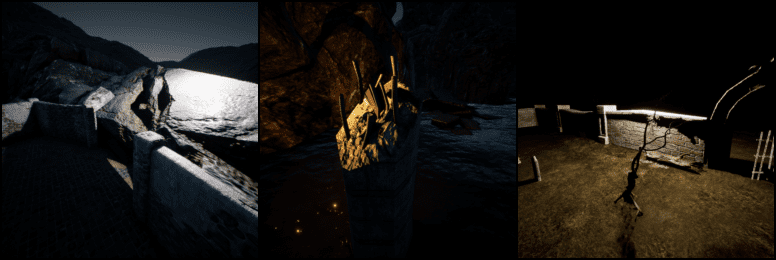}
    \end{minipage} \hfill
    \begin{minipage}[c]{0.49\linewidth}
        \includegraphics[width=1\textwidth]{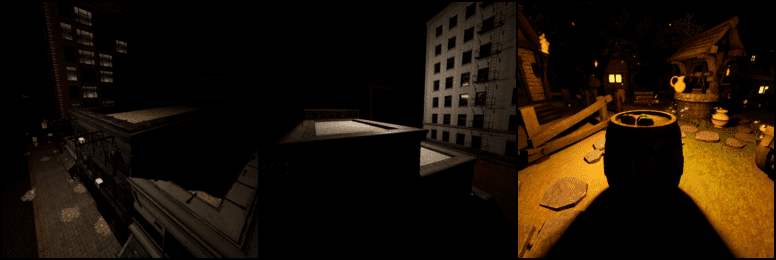}
    \end{minipage}
    \begin{minipage}[c]{0.49\linewidth}
        \includegraphics[width=1\textwidth]{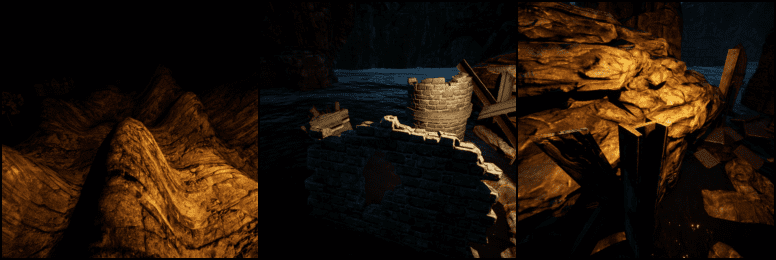}
    \end{minipage} \hfill
    \begin{minipage}[c]{0.49\linewidth}
        \includegraphics[width=1\textwidth]{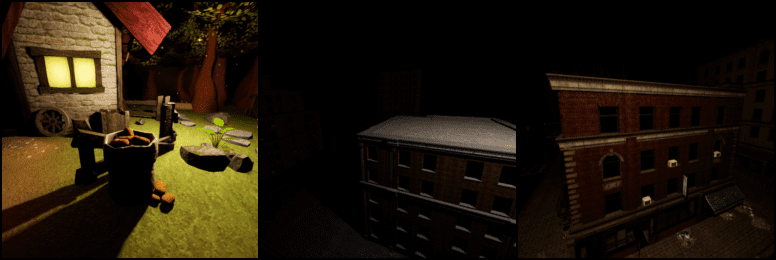}
    \end{minipage}
    \begin{minipage}[c]{0.49\linewidth}
        \includegraphics[width=1\textwidth]{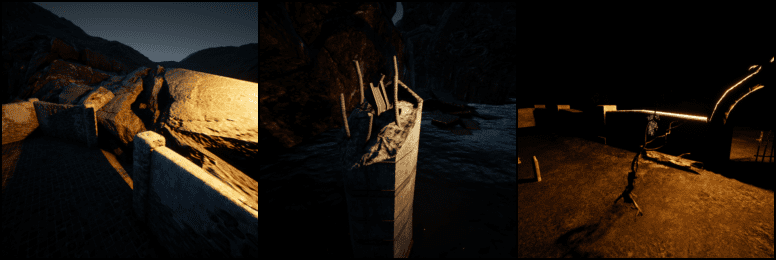}
    \end{minipage} \hfill
    \begin{minipage}[c]{0.49\linewidth}
        \includegraphics[width=1\textwidth]{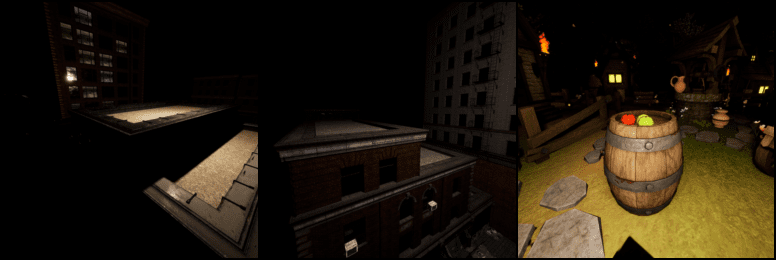}
    \end{minipage}
    \begin{minipage}[c]{0.49\linewidth}
        \includegraphics[width=1\textwidth]{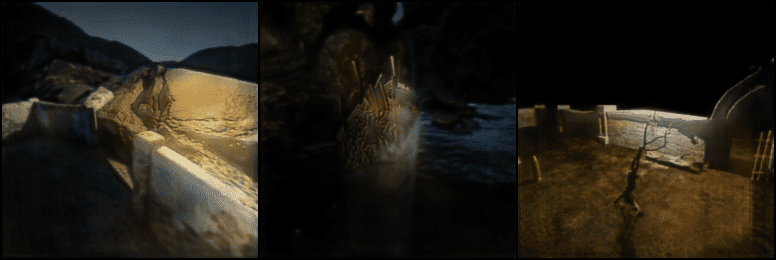}
    \end{minipage} \hfill
    \begin{minipage}[c]{0.49\linewidth}
        \includegraphics[width=1\textwidth]{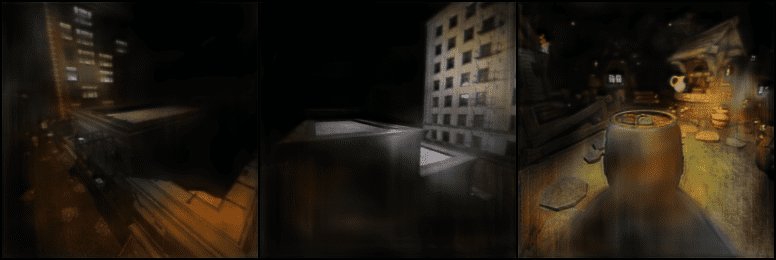}
    \end{minipage}
    \caption{Example of relighting performed. From top to bottom: input $I$, target $T$ and ground-truth $G(I,T)$ images; relit image $\hat{G}(I,T)$ produced by the model with illumination predicter without discriminator. From left to right: 3 samples from the train set, 3 samples from the evaluation set.}
    \label{fig:resultsnogan}
\end{figure}

\begin{figure}[H]
    \centering
    \begin{minipage}[c]{0.49\linewidth}
        \includegraphics[width=1\textwidth]{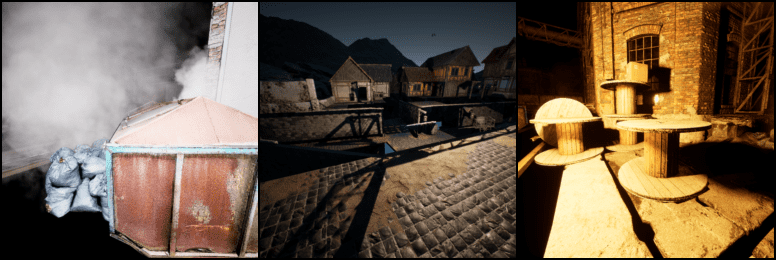}
    \end{minipage} \hfill
    \begin{minipage}[c]{0.49\linewidth}
        \includegraphics[width=1\textwidth]{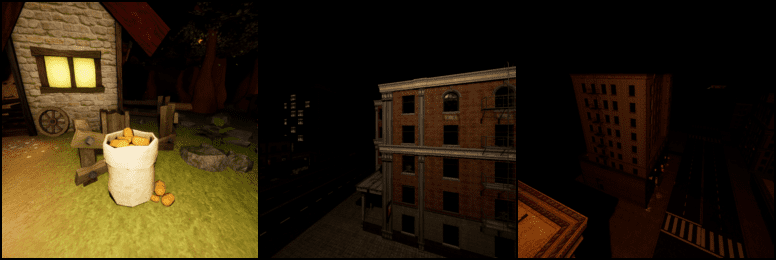}
    \end{minipage}
    \begin{minipage}[c]{0.49\linewidth}
        \includegraphics[width=1\textwidth]{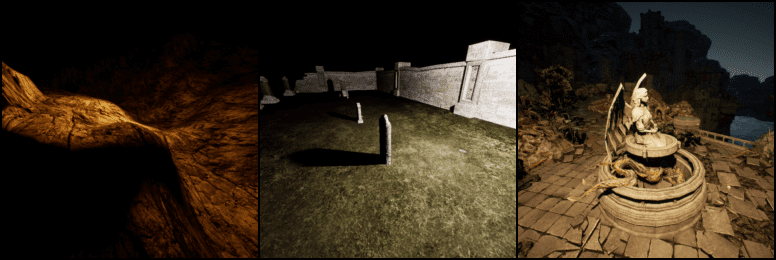}
    \end{minipage} \hfill
    \begin{minipage}[c]{0.49\linewidth}
        \includegraphics[width=1\textwidth]{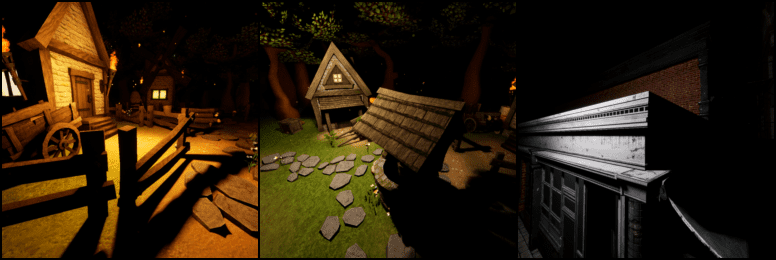}
    \end{minipage}
    \begin{minipage}[c]{0.49\linewidth}
        \includegraphics[width=1\textwidth]{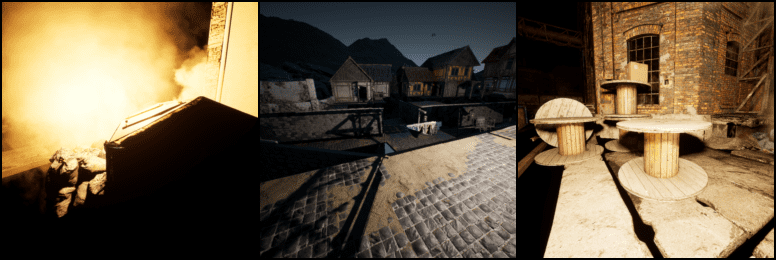}
    \end{minipage} \hfill
    \begin{minipage}[c]{0.49\linewidth}
        \includegraphics[width=1\textwidth]{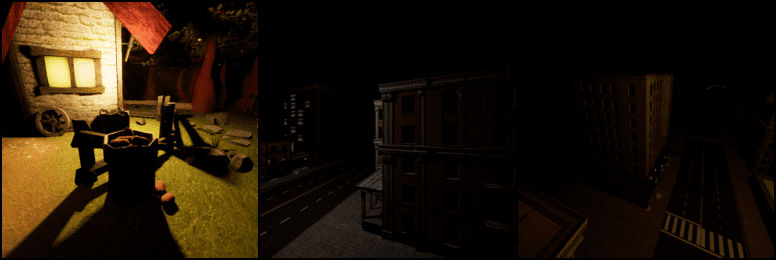}
    \end{minipage}
    \begin{minipage}[c]{0.49\linewidth}
        \includegraphics[width=1\textwidth]{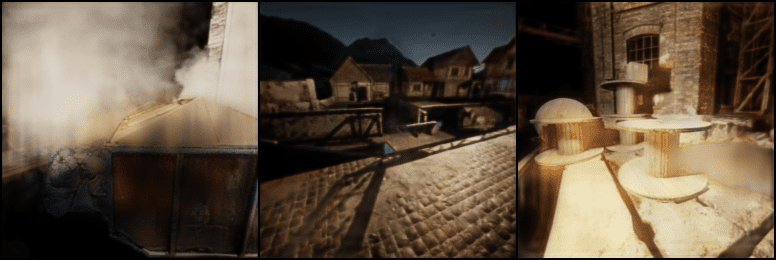}
    \end{minipage} \hfill
    \begin{minipage}[c]{0.49\linewidth}
        \includegraphics[width=1\textwidth]{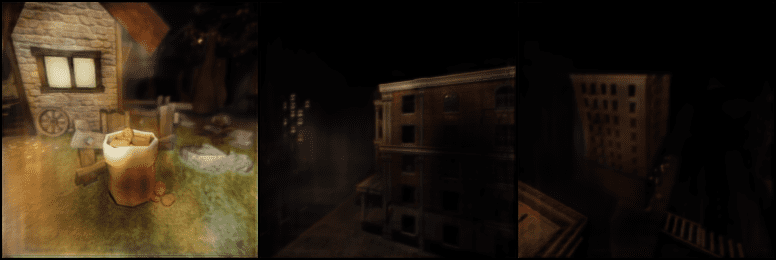}
    \end{minipage}
    \caption{Example of relighting performed. From top to bottom: input $I$, target $T$ and ground-truth $G(I,T)$ images; relit image $\hat{G}(I,T)$ produced by the model with illumination predicter with discriminator. From left to right: 3 samples from the train set, 3 samples from the evaluation set. By mistake, we used $\ell_2$ instead of the cosine-based loss $\mathcal{L}_{d}$ for this experiment.}
    \label{fig:resultsgan}
\end{figure}

\section{Environment maps estimation}

\begin{figure}[H]
    \centering
    \begin{minipage}[c]{0.49\linewidth}
        \includegraphics[width=1\textwidth]{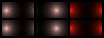}
    \end{minipage} \hfill
    \begin{minipage}[c]{0.49\linewidth}
        \includegraphics[width=1\textwidth]{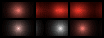}
    \end{minipage}
    \begin{minipage}[c]{0.49\linewidth}
        \includegraphics[width=1\textwidth]{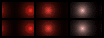}
    \end{minipage} \hfill
    \begin{minipage}[c]{0.49\linewidth}
        \includegraphics[width=1\textwidth]{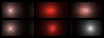}
    \end{minipage}
    \caption{Environment map predictions and generated ground-truths for the model with only light representation in the latent variable. Presented results are from the 9\textsuperscript{th} epoch of training (train environment maps; left column of the figure) and after 8 epochs of training (test environment maps; right column of the figure). Each of four sets of three environment maps consists of predictions by the network (top row) and generated ground-truths that were guiding the prediction through $\mathcal{L}_e$ loss in equation \ref{eq:envmapLoss} (bottom row). Two top sets relate to environment maps for the input images, two bottom sets -- to environment maps for target images.}
    \label{fig:envmaps}
\end{figure}

\begin{figure}[H]
    \centering
    \begin{minipage}[c]{0.49\linewidth}
        \includegraphics[width=1\textwidth]{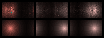}
    \end{minipage} \hfill
    \begin{minipage}[c]{0.49\linewidth}
        \includegraphics[width=1\textwidth]{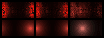}
    \end{minipage}
    \begin{minipage}[c]{0.49\linewidth}
        \includegraphics[width=1\textwidth]{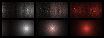}
    \end{minipage} \hfill
    \begin{minipage}[c]{0.49\linewidth}
        \includegraphics[width=1\textwidth]{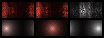}
    \end{minipage}
    \caption{Environment map predictions and generated ground-truths for the model with light and scene representation in the latent variable. Presented results are from the 10\textsuperscript{th} epoch of training (train environment maps; left column of the figure) and after 9 epochs of training (test environment maps; right column of the figure). Each of four sets of three environment maps consists of predictions by the network (top row) and generated ground-truths that were guiding the prediction through $\mathcal{L}_e$ loss in equation \ref{eq:envmapLoss} (bottom row). Two top sets relate to environment maps for the input images, two bottom sets -- to environment maps for target images.}
    \label{fig:envmapsScene}
\end{figure}

\section{First experiments to familiarize with the dataset}

\begin{figure}[H]
    \centering
    \includegraphics[width=0.40\textwidth]{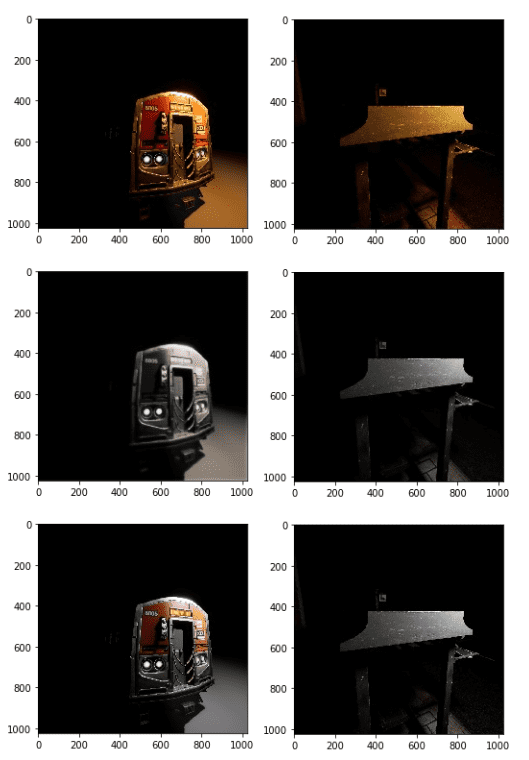}
    \caption{1-to-1 relighting of $1024 \times 1024$ images $I$ with $c_I = 2500\si{\kelvin}$ and $d_I = E$ to images $\hat G \approx G$ with $c_G = 6500\si{\kelvin}$ and $d_G = E$. Left is after 2 epochs of training, right after 20 epochs. Top is $1024 \times 1024$ images from the evaluation set $I$ with $c_I = 2500\si{\kelvin}$ and $d_I = E$, bottom is $1024 \times 1024$ images from the evaluation set $G$ with $c_G = 6500\si{\kelvin}$, $d_G = E$ and $S_G = S_I$, middle is relit image $\hat G$ produced by the network.}
    \label{fig:beginningExperiment1to1}
\end{figure}

Among our first experiments to habituate to the dataset, we slightly modified the depth estimation network used in \cite{depth} to have 3 channels as output instead of 1 and trained it with ground-truth images $G$ to perform 1-to-1 relighting of $1024 \times 1024$ images $I$ with $c_I = 2500\si{\kelvin}$ and $d_I = E$ to images $\hat G \approx G$ with $c_G = 6500\si{\kelvin}$, $d_G = E$ and $S_G = S_I$, as depicted in the figure \ref{fig:beginningExperiment1to1}.

We also experimented with the trained network of \cite{deepportrait} to relight images of portraits as well a non-portrait image. As expected, we observed good results on portrait images -- as long as the head is not too small, too far to the camera -- and bad results on the non-portrait image where the shadows appearing on the relit images seem to reveal a non-existing face.

%\subsection{Difficult interpretability of the light latent variables without ground-truth envmaps }

\section{Code organization}

The code can be found in our repository: \url{https://github.com/martin-ev/2DSceneRelighting} and is divided into three core parts: 
\begin{enumerate}
    \item \textbf{Training scripts} containing entire training logic, such as data loading, model setup, training progress monitoring with Tensorboard, main training and evaluation loops. These scripts can be found in the main folder of the project: \texttt{train.py} and \texttt{trainWithGeneratedEnvmaps.py}. The former allows for training the networks with illumination predictor with modifiable training parameters for which the configurations can be found in the \texttt{configurations} folder. The \texttt{evaluate.py} script was used to compute the metrics for the evaluated models presented in table \ref{table:results}.
    \item \textbf{Model architecture} components in \texttt{models} module. All the components for the encoder, decoder and latent representation manipulations can be found in \texttt{models/swapModels.py}. The \texttt{models/patchGan.py} contains the PathGAN discriminator implementation (taken from the code linked to \cite{img2img}).
    \texttt{models/old} contain various code used in early experiments of the project including the implementation of architecture from \cite{portrait} that was later generalized and put into \texttt{models/swapModels.py}.
    \item \textbf{Utility functions} module called \texttt{utils}. It contains: dataset loading classes (\texttt{dataset.py}), CUDA device setup (\texttt{device.py}), environment map generation code (\texttt{envmap.py}), implementation of various losses (\texttt{losses.py}), helper functions wrapping calls to PSNR and SSIM metrics from Kornia library \cite{kornia} (\texttt{metrics.py}), functions for storing and loading the saved models (\texttt{storage.py}), tools to start, stop and interact with the Tensorboard (\texttt{tensorboard.py}).
\end{enumerate}

The old experiments and training code are located in \texttt{misc/oldExperiments} and \texttt{misc/others} contains Jupyter notebooks with tests of some parts of the implementation. \texttt{lpips\_pytorch} contains the implementation of LPIPS we used and that comes from \cite{lpipsCode}.

\end{document}